%% file: main.tex
\documentclass[10pt,twocolumn,letterpaper]{article}

\usepackage[pagenumbers]{cvpr} % To force page numbers, e.g. for an arXiv version

\input{preamble}

\definecolor{cvprblue}{rgb}{0.21,0.49,0.74}
\usepackage[pagebackref,breaklinks,colorlinks,allcolors=cvprblue]{hyperref}

\def\paperID{} % *** Enter the Paper ID here
\def\confName{ECV at CVPR}
\def\confYear{2026}

\title{LILogic Net: Compact Logic Gate Networks with Learnable Connectivity \\for Efficient Hardware Deployment}

\author{
Katarzyna Fojcik$^{1}$ \quad
Renaldas Zioma$^{2}$ \quad
Jogundas Armaitis$^{2}$\\
$^{1}$Wroclaw University of Science and Technology \quad
$^{2}$Transcendent Logic \\
{\tt\small katarzyna.fojcik@pwr.edu.pl} \quad
{\tt\small rz@TranscendentLogic.com} \quad
{\tt\small ja@TranscendentLogic.com}\\
}
\begin{document}
\maketitle
\input{sec/0_abstract}   
\input{sec/1_introduction}
\input{sec/2_related_work}

\input{sec/3_method}
\input{sec/4_experiments}

\input{sec/5_conclusion}

{
    \small
    \bibliographystyle{ieeenat_fullname}
    \bibliography{main}
}

\input{sec/X_suppl}

\end{document}

%% file: preamble.tex
\usepackage{microtype}

\renewcommand{\paragraph}[1]{\vspace{.5em}\noindent\textbf{#1.}}

\usepackage{graphicx}
\usepackage{amsmath}
\usepackage{amssymb}
\usepackage{booktabs}
\usepackage{multirow}
\usepackage{subcaption}
\usepackage{makecell}
\usepackage{placeins}
\usepackage{float}

%% file: sec/0_abstract.tex
\begin{abstract}

Efficient machine learning deployment requires models that account for hardware constraints. Because binary logic gates are the fundamental primitives of digital hardware, models built directly from logic operations offer a promising path toward highly energy-efficient computation. Recent work has shown that networks of binary logic gates can be trained with gradient-based optimization and that their wiring can be learned. However, existing approaches remain limited in scalability and training efficiency. We address these challenges by treating the network connectome as a differentiable object and introducing a Top-K connectivity mechanism that enforces structured sparsity during training. Our resulting architecture, LILogicNet, substantially improves the efficiency of logic-gate networks. A model with only 8,000 gates trains on MNIST in under five minutes while achieving 98.45\% test accuracy, matching the performance of state-of-the-art logic-gate models that require two orders of magnitude more gates. At larger scales, a 256,000-gate model achieves 60.98\% test accuracy on CIFAR-10, surpassing prior approaches with comparable gate budgets. Because the final model is fully binarized and composed entirely of logic operations, inference incurs minimal compute overhead and maps naturally to a wide range of digital hardware platforms, enabling efficient deployment across diverse computing systems.
\end{abstract}

%% file: sec/1_introduction.tex
\section{Introduction}
\label{sec:intro}

In recent years, scaling up machine learning models has driven major advances across real-world applications, enabled by specialized datacenter hardware~\cite{jouppi2017datacenter}. In such environments, hardware cost and energy consumption are typically secondary concerns. As machine learning matures, however, there is growing interest in deploying models on low-power, cost-sensitive edge devices~\cite{han2015deep}, demanding tighter co-design between model architecture and hardware capabilities to extend AI beyond traditional datacenter settings~\cite{banbury2021mlperf}. Modern machine learning models largely rely on matrix multiplication operations, efficiently supported by high-level APIs and massively parallel hardware such as GPUs~\cite{sze2017efficient,geng2024efficient}. Computational hardware fundamentally operates on binary logical gate operations, which are directly available across ASICs, FPGAs~\cite{kuon2007fpga,moolchandani2022review}, CPUs, and microcontrollers~\cite{champs2023survey,silvano2023survey,luo2024embedded}. Designing models that natively operate on binary signals and logical gates, therefore, represents a promising direction for hardware-efficient AI.

\begin{figure}
    \centering
    \includegraphics[width=\linewidth]{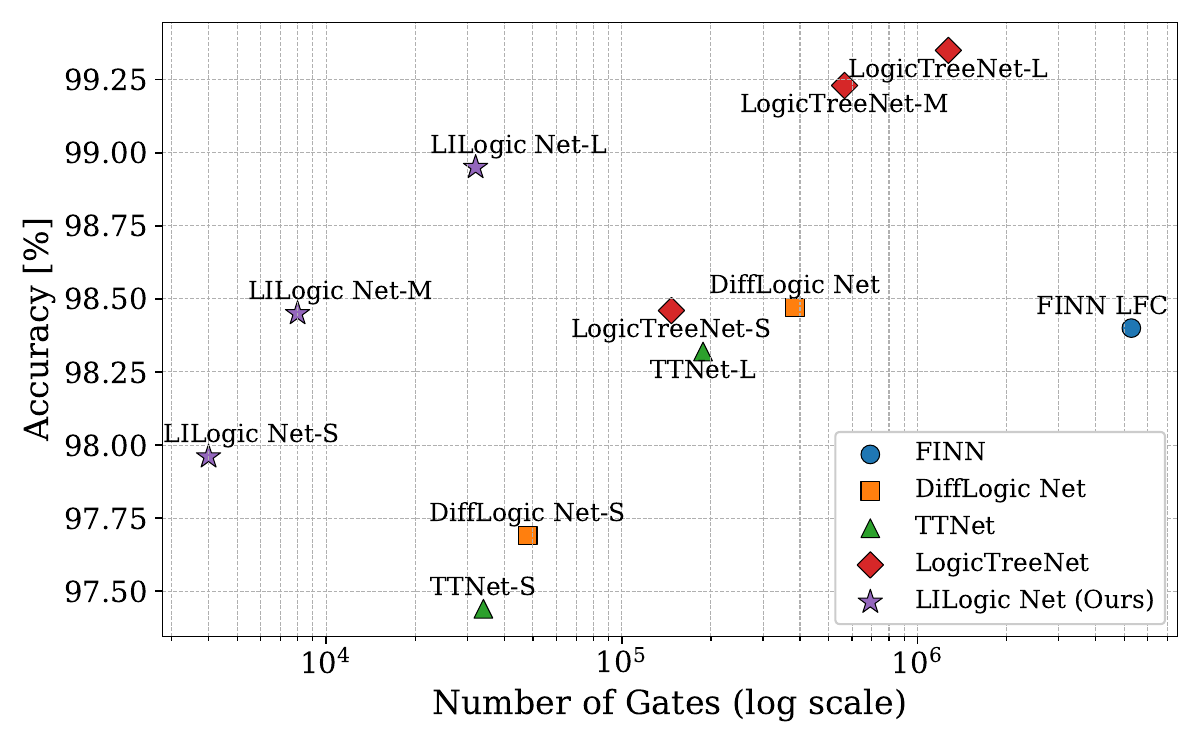}
    \caption{Gate count vs.\ accuracy on the MNIST dataset. Our models LILogic Nets achieve significantly higher accuracy for a given gate budget compared to state-of-the-art baselines, lying well above the Pareto front.}
    \label{fig:comparison_sota}
\end{figure}

One approach is to replace components of state-of-the-art architectures, such as multilayer perceptrons (MLPs)~\cite{goodfellow2016deep} and convolution layers~\cite{lecun1998gradient,zhao2024review}, with logical counterparts. Logical Gate Networks (LGNs), composed of randomly wired layers of trainable two-input logical gates, have shown that purely logical architectures can be trained end-to-end using gradient descent~\cite{petersen2022deep}. On benchmarks such as MNIST and CIFAR-10, these models achieved competitive accuracy, though typically with large gate budgets and fixed connectomes. Subsequent work demonstrated that wiring itself can be learned. In particular, Bacellar et al.~\cite{bacellar2024differentiable} introduced Differentiable Wiring Networks (DWN), where discrete interconnections between lookup tables are optimized via softmax relaxation and discretized at inference. Other studies explored learnable connectomes primarily in tabular settings and from an interpretability perspective~\cite{yue2024learning,yue2025learning}, but were limited to relatively small networks—up to 2,000 gates—citing training slowdowns due to computational burden.

If LGNs are to serve as practical replacements for MLP components, it is crucial not only to learn connectivity but also to do so efficiently and at scale. In this work, we treat the connectome as a differentiable object and explicitly target scalability and structured sparsity. We introduce a Top-K connectivity mechanism that enforces sparsity during training rather than through post-hoc pruning, improving optimization stability and reducing memory footprint. Furthermore, we propose a Basis Projection (BasisProj) method for efficient gradient-based computation of logic gate outputs.

We introduce \textbf{LILogic Net} (\emph{Learnable Interconnect Logic Network}), a compact and scalable LGN architecture achieving 98.95\% accuracy on MNIST using only 8,000 logic gates and 60.98\% on CIFAR-10 at larger gate budgets. Compared to prior LGN approaches, 
LILogic Net achieves improved accuracy at comparable or smaller gate counts, demonstrating that (sparse) connectome learning yields measurable end-to-end gains. Code will be made publicly available at \url{https://github.com/jogundas/LILogicNet}.

We summarize our contributions as follows:

% To facilitate further research, our framework also enables models with a mix of fixed and learnable connectome layers. 

% To facilitate further research, we will release the full source code of our framework (upon acceptance), which also enables models with a mix of fixed and learnable connectome layers.

\begin{itemize}

\item \textbf{Top-K connectivity}: A mechanism that enforces sparsity during training by selecting the strongest connection among K random candidates per gate.
\item \textbf{Basis Projection (BasisProj)}: Efficiently computes gate outputs via a fixed projection, yielding up to 4× training speedup.
\item \textbf{LILogicNet architecture}: Compact logical gate networks that achieve competitive or better accuracy on MNIST, FashionMNIST, and CIFAR-10 with up to two orders of magnitude fewer gates than prior models.
\end{itemize}

% Our results indicate that LGNs with learnable and structured connectivity can scale effectively while preserving hardware-native computation, highlighting their potential as alternatives to matrix-multiplication-based models in resource-constrained environments.

%% file: sec/2_related_work.tex
\section{Related Work}

Several key trends have emerged in binarized and hardware-efficient neural models. Binary Neural Networks (BNNs)~\cite{yuan2023comprehensive}, such as XNOR-Net~\cite{rastegari2016xnor} and BinaryNet~\cite{hubara2016binarized}, showed that binary weights and activations can significantly reduce memory and computation costs with minimal accuracy loss. In parallel, Quantized Neural Networks~\cite{hubara2018quantized} reinforced the promise of low-precision optimization. IR-Net~\cite{qin2020forward} addressed information loss in binary training via forward–backward techniques like Libra Parameter Binarization and Error Decay Estimators, while ReActNet~\cite{liu2020reactnet} improved binary model accuracy through generalized activations. From a deployment perspective, FINN~\cite{umuroglu2017finn} introduced a scalable FPGA framework for real-time BNN inference.

% A closely related hardware-aware direction explores LUT-based computation, which directly exploits FPGA primitives for neural representations. LUTNet (Wang et al., 2019) replaced binary exclusive NOR (XNOR) operations in BNNs with arbitrary K-input LUTs to increase logic density and enable aggressive pruning tailored to FPGA resources. LogicNets (Umuroglu et al., 2020) extended this idea by co-designing sparse, quantized networks where fan-in–limited neurons map directly to truth tables, enabling deeply pipelined circuits for extreme-throughput applications. More recently, Differentiable Weightless Neural Networks, generalizing classical weightless and LUT-based models, learn both address mapping and reduction operations over symbolic inputs using hybrid differentiable/combinatorial optimization (Bacellar et al., 2025). For a comprehensive overview, Guo (2025) surveys LUT-based FPGA DNNs, covering training schemes from gradient-based to combinatorial and hybrid approaches.
A parallel line of research investigates neural architectures designed explicitly around FPGA lookup tables (LUTs) as the primary computational primitive. Rather than mapping conventional neural layers onto hardware post hoc, these approaches restructure computation to better match the native logic fabric. LUTNet~\cite{wang2019lutnet} was an early example, using learnable K-input lookup tables to improve logic utilization and enable structured pruning aligned with FPGA resources. LogicNets~\cite{umuroglu2020logicnets} pushed this further and increased throughput by constraining neuron fan-in so that each unit could be realized directly as a truth table. More recent developments extend this idea beyond binarized weight abstractions. DWNs~\cite{bacellar2024differentiable} revisit classical weightless models and introduce learnable address mappings and aggregation mechanisms over symbolic inputs, combining gradient-based training with combinatorial components. 
% A broader perspective on LUT-centric deep learning methods, including gradient-based, hybrid, and non-differentiable optimization strategies, is provided in the survey by Guo (2025).

% A distinct but conceptually related direction stems from differentiable logic-based neural architectures, which structure computation around learnable logic gates rather than weighted connections. Logic Gate Networks (LGNs), introduced by Petersen et al.~\cite{petersen2022deep}, demonstrated that logic gate arrays--traditionally considered incompatible with gradient descent--can be relaxed and trained using standard deep learning methods. Even in their early form, LGNs offered benefits over BNNs in terms of architectural simplicity, interpretability, and hardware alignment. Subsequent work explored both theoretical and practical aspects of LGNs; Yue and Jha investigated interpretability in tabular data using straight-through estimators, Gumbel-softmax, and preprocessing schemes~\cite{yue2024learning,yue2025learning}. However, these early LGNs were limited in size and not performance-optimized.

A related direction is differentiable logic-based neural architectures, which replace weighted connections with learnable logic gates. Logic Gate Networks (LGNs), introduced by Petersen et al.~\cite{petersen2022deep}, showed that such structures can be relaxed and trained with gradient descent. Even early LGNs offered advantages over BNNs in simplicity, interpretability, and hardware alignment. Later work explored their theoretical and practical aspects, including interpretability methods~\cite{yue2024learning,yue2025learning}, though these models remained limited in scale and performance.

Recent advances have addressed these limitations. Yousefi et al.~\cite{yousefi2025mind} improved training stability and gate utilization by mitigating the discretization gap using Gumbel noise and Hessian regularization. Petersen et al.~\cite{petersen2024convolutional} proposed TreeLogicNet, incorporating logic gates into convolutional windows with logical $OR$ pooling, achieving 99.35\% accuracy on MNIST and 86.29\% on CIFAR-10—but using 2–4 orders of magnitude more gates than our method. Although based on logical operations, TreeLogicNet retains CNN-like receptive fields and parameter sharing, resulting in higher gate and operation counts. Other contributions include the eXpLogic framework~\cite{wormald2025explogic}, which enhances interpretability through saliency maps and gate pruning with minimal accuracy drop. Kresse et al.~\cite{kresse2025logic} showed that LGNs are inherently suitable for formal verification using SAT solvers, enabling proofs of fairness and robustness, and Clester~\cite{clester2025synthesizing} explored LGNs for generative applications such as music synthesis. On the hardware side, Wang et al.~\cite{wang2025logic} introduced four custom RISC-V instructions to accelerate LGN inference, achieving 95.8\% accuracy on MNIST with a compact ~18 K logic gate network and reduced runtime by over 87\% compared to a standard RV32IM core.

%% file: sec/3_method.tex
\section{Method}

\subsection{Differentiable Logical Gate Networks}

\begin{figure*}[t]
    \centering
    \begin{subfigure}[t]{0.30\textwidth}
        \centering
        \includegraphics[width=\linewidth]{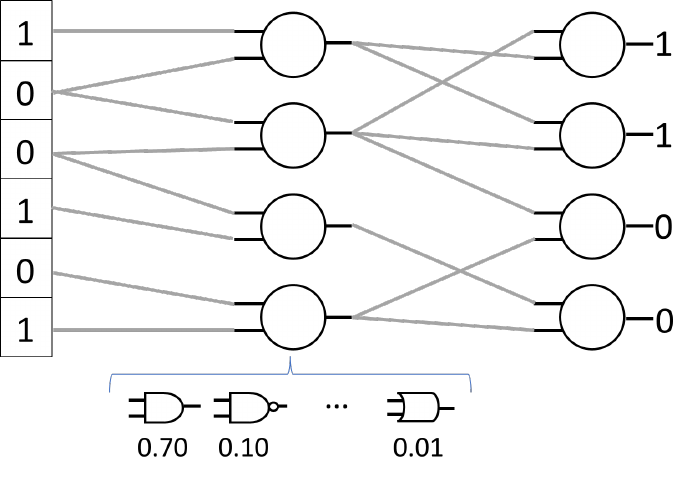}
        \caption{Fixed Connectome}
        \label{fig:a}
    \end{subfigure}
    \hfill
    \begin{subfigure}[t]{0.30\textwidth}
        \centering
        \includegraphics[width=\linewidth]{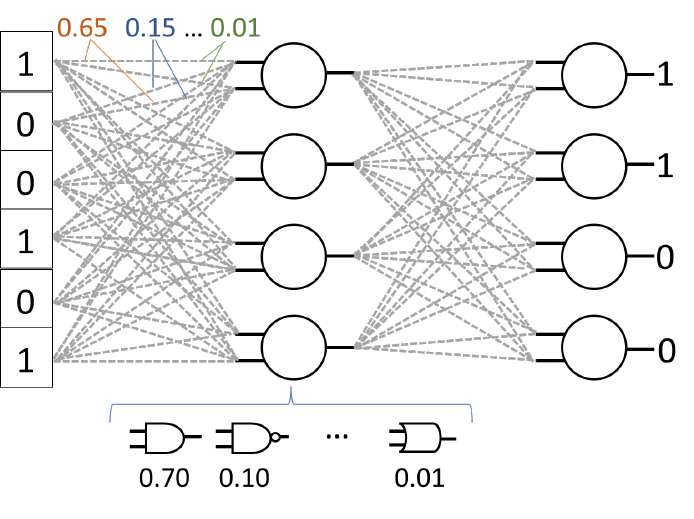}
        \caption{Fully Learnable Dense Connectome}
        \label{fig:c}
    \end{subfigure}
    \hfill
    \begin{subfigure}[t]{0.30\textwidth}
        \centering
        \includegraphics[width=\linewidth]{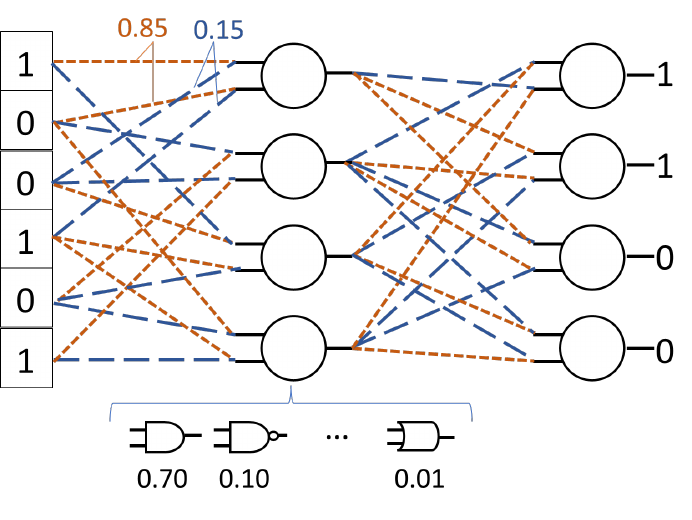}
        \caption{Top-K Sparse Connectome (here K=2)}
        \label{fig:b}
    \end{subfigure}
    \caption{Overview of training-time interconnect strategies. Each node learns a distribution over the 16 Boolean functions. Input connections are either fixed or modeled as learnable distributions, depending on the connectome design.}
    \label{fig:training_models}
\end{figure*}

\begin{figure}
    \centering
    \includegraphics[width=\linewidth]{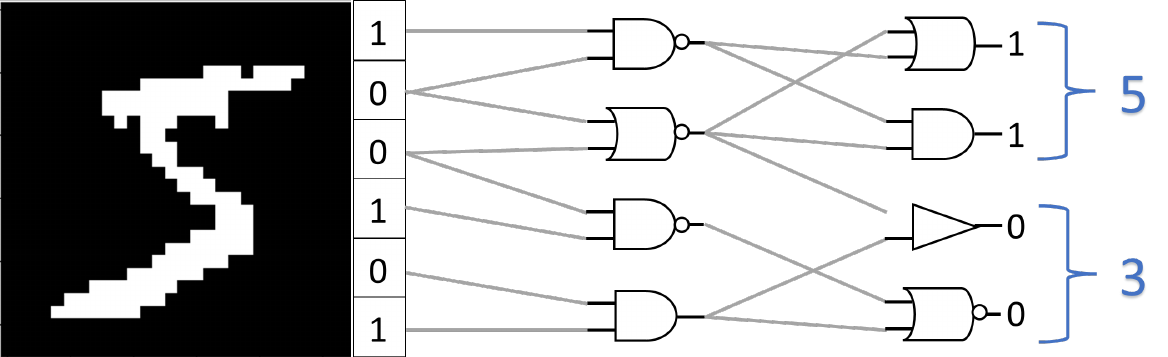}
    \caption{Inference pipeline: all logic gates and connections are fully binarized, yielding a compact, deterministic circuit.}
    \label{fig:inference_model}
\end{figure}

Traditional Logic Gate Networks (LGNs) are sparse, weightless models built from binary logic gates (e.g., AND, XOR). While highly efficient for inference, their discrete nature prevents gradient-based training, thus limiting scalability. The Differentiable LGN framework introduced by Petersen et al.~\cite{petersen2022deep} solves this by: (i) relaxing binary inputs to continuous values in $[0,1]$,
(ii) using probabilistic interpretations of logic gates (e.g., $A \wedge B \approx A \cdot B$~\cite{zadeh1965fuzzy,van2022analyzing}), and (iii) representing each node as a soft combination over all 16 2-input Boolean functions, enabling learning via gradient descent.

This makes LGNs trainable with standard optimizers (e.g., Adam~\cite{kingma2015adam}), and—after training—easily discretizable to efficient, purely combinatorial LGNs, which means that they have no "memory" or "state".

We design an LGN architecture that jointly learns the logic gate behavior and the structure of connections between them (the connectome), enabling both efficient training and inference. Our key methodological contributions are twofold: (i) a more GPU-efficient gate representation, and (ii) a systematic comparison of several strategies for interconnect optimization.

\subsection{Model Architecture} \label{sec:architecture}

Our models have 1–4 logic gate layers, each containing a fixed number of gates selected from $\{2\mathrm{K}, 4\mathrm{K}, 8\mathrm{K}, 16\mathrm{K}, 32\mathrm{K}, 64\mathrm{K}, 128\mathrm{K}\}$. Each node receives two relaxed binary inputs and implements one of the 16 Boolean functions~\cite{roth2004fundamentals}. All layers share the same width and interconnect strategy. We consider three strategies for wiring between layers (Figure~\ref{fig:training_models}):

\begin{itemize}
    % \item \textbf{(F) Fixed Connectome:} A static, non-learnable connection scheme where each gate receives its inputs from a random subset of the previous layer. This setting mirrors the original Differentiable LGN formulation~\cite{petersen2022deep} and serves as our baseline. While universal, its lack of connectome specialization typically requires deeper networks with wider layers to achieve competitive accuracy.
    % \item \textbf{(L) Fully learnable dense connectome:} A differentiable connection matrix parameterized via softmax over all possible predecessor nodes. Each gate independently samples two inputs from this full set without replacement. This approach is expressive but introduces higher memory and compute cost, especially for wider layers.
    % \item \textbf{(Top-K) Top-K Sparse Connectome:} Each output gate is assigned a limited number of input candidates $K$, selected at initialization~\cite{han2015deep}. Only the connection weights within this subset are optimized during training. We experiment with a wide range of $K$ values: $\{2, 4, 8, 16, 32, 64, 128\}$, enabling a trade-off between sparsity and expressiveness.
    \item \textbf{(F) Fixed connectome:} Each gate receives inputs from a random subset of the previous layer. This mirrors the original Differentiable LGN (DiffLogicNet)~\cite{petersen2022deep} and serves as a baseline, typically requiring deeper or wider layers to match accuracy.
    \item \textbf{(L) Fully learnable dense connectome:} For each node, both of its two inputs are connected to all outputs of the previous layer (or to the input vector for the first layer). The connection weights for each input are parameterized separately via a differentiable softmax.
    \item \textbf{(Top-K) Sparse connectome:} For each node, both of its two inputs independently select $K$ random candidate connections from the previous layer (or the input vector for the first layer) at initialization, with each candidate assigned a learnable logit sampled from a standard normal distribution. During training, a softmax over the $K$ candidates produces a weighted combination for each input. We experiment with $K \in \{2,4,8,16,32,64,128\}$ to trade off sparsity and expressiveness.

\end{itemize}
% \paragraph{(F) Fixed Connectome:} A static, non-learnable connection scheme where each gate receives its inputs from a random subset of the previous layer. This setting mirrors the original Differentiable LGN formulation~\cite{petersen2022deep} and serves as our baseline. While universal, its lack of connectome specialization typically requires deeper networks with wider layers to achieve competitive accuracy.
% \paragraph{(L) Fully Learnable Dense Connectome:} A differentiable connection matrix parameterized by softmax, enabling each gate to learn input pairings directly from data. Though expressive, this approach introduces high memory and compute cost during  training, especially for wider layers. 
% \paragraph{(Top-K) Top-K Sparse Connectome:} Each output gate is assigned a limited number of input candidates $K$, selected at initialization~\cite{han2015deep}. Only the connection weights within this subset are optimized during training. We experiment with a wide range of $K$ values: $\{2, 4, 8, 16, 32, 64, 128\}$, enabling a trade-off between sparsity and expressiveness.
\noindent
After training, all models are fully binarized. Each gate selects its most probable logic gate and inputs (for $L$ and Top-$K$ connectomes), yielding a fixed-size, deterministic binary circuit suitable for efficient inference (Figure~\ref{fig:inference_model}).

\subsection{Dataset and Preprocessing}

% We evaluated our models on the MNIST dataset, which consists of $60{,}000$ training and $10{,}000$ test images of handwritten digits~\cite{lecun1998mnist}. Each grayscale $28 \times 28$ image was flattened into a 784-dimensional input vector, and binarized by thresholding pixel intensities at a fixed threshold of $0.25$.
We evaluated our models on the MNIST, FashionMNIST, and CIFAR-10 datasets.
The MNIST dataset consists of $60{,}000$ training and $10{,}000$ test images of handwritten digits~\cite{lecun1998mnist}. Each grayscale image of size $28 \times 28$ was flattened into a 784-dimensional input vector and binarized using a fixed intensity threshold of $0.25$.
The FashionMNIST dataset contains $60{,}000$ training and $10{,}000$ test grayscale images of fashion items across ten categories~\cite{xiao2017fashion}. Each $28 \times 28$ image was flattened and binarized using fixed thresholds $[0.125,, 0.25,, 0.375,, 0.5,, 0.625,, 0.75,, 0.875]$, resulting in an effective input dimensionality of $5{,}488$.
The CIFAR-10 dataset contains $50{,}000$ training and $10{,}000$ test color images of $32 \times 32$ pixels across ten object categories~\cite{krizhevsky2009learning}. For CIFAR-10, each RGB channel was processed independently and binarized using the same thresholds $[0.125,, 0.25,, 0.375,, 0.5,, 0.625,, 0.75,, 0.875]$. The resulting binary maps from all channels were concatenated into a single flattened vector, yielding an effective input dimensionality of $21{,}504$.

% To improve generalization~\cite{simard2003best}, we applied a diverse set of augmentations to the training data using the \texttt{torchvision.transforms.v2} interface. These included random rotation ($\pm 10^\circ$), shear (up to $10^\circ$), scaling (by $\pm 10\%$), and elastic deformations with $\alpha = 64$ and $\sigma = 6$. Each augmented transformation was applied independently to generate $10$ distinct copies of the training dataset, leading to a $10 \times$ expanded training set. 
Learning interconnections increases model capacity, which can lead to overfitting in low-data regimes without augmentation. We applied dataset-specific transformations to improve generalization~\cite{simard2003best}.
For MNIST, we used the \texttt{torchvision.transforms.v2} interface to apply random rotation ($\pm 10^\circ$), shear (up to $10^\circ$), scaling (by $\pm 10\%$), and elastic deformations with $\alpha = 64$ and $\sigma = 6$. Each augmented transformation was applied independently to generate $10$ distinct copies of the training dataset, leading to a $10 \times$ expanded training set.
For FashionMNIST, no augmentations were applied.
For CIFAR-10, we applied standard natural image augmentations~\cite{shorten2019survey}, including random cropping to $32 \times 32$ with 4-pixel reflection padding and random horizontal flips with 50\% probability, expanding the training set $8 \times$.
We note that the original DiffLogicNet baseline~\cite{petersen2022deep} does not use augmentations. To enable a fair comparison and isolate the effect of learnable connectivity, we additionally perform evaluations of our models against versions with fixed connectomes under the same augmentation settings.

The validation set was constructed by applying only resizing and binarization (i.e., no augmentation) to the original training images, and the same preprocessing was used for the test set. A fixed random seed ensured reproducibility of the dataset splits and augmentation pipeline.

\subsection{Training and Inference}

All models were trained for 200 epochs using the Adam optimizer~\cite{kingma2015adam} with a learning rate of 0.075, a batch size of 256, and no additional regularization such as dropout or weight decay. Our experiments focus purely on the architectural and interconnect design choices, especially different sparsity regimes, and aim to isolate their effect without the influence of external regularizers. Top-$K$ interconnect variants act implicitly as a form of regularization by constraining the connectivity space.

% \noindent
% \subsubsection{Timing measurements} Reported training times include the full training procedure with validation but exclude the initial data loading and the online generation of augmented samples. This ensures that the reported times reflect only the model's forward and backward passes during training.

\subsubsection{Differentiable Relaxation}

To enable gradient-based training of logic gate selection, each node is 
parameterized by a vector of 16 real-valued weights, $\mathbf{w} \in \mathbb{R}^{16}$, 
representing logits over the space of all 16 possible binary logic gates. 
These logits are transformed into a probability distribution via the  softmax function, possibly scaled by a temperature parameter $\tau_g$~\cite{agarwala2020temperature}, 
which we kept fixed at $\tau_g = 1$ in all our experiments:
\begin{equation}
\mathbf{p} = \text{softmax}(\tau_g \cdot \mathbf{w}), \quad
p_i = \frac{\exp(\tau_g \cdot w_i)}{\sum_j \exp(\tau_g \cdot w_j)}.
\end{equation}

The resulting probability vector $\mathbf{p} \in \Delta^{15}$ lies in the 
15-dimensional probability simplex, meaning that all entries are non-negative 
and sum to 1: $\sum_{i=1}^{16} p_i = 1$, $p_i \geq 0$. This formulation treats 
gate selection as a categorical distribution that can be trained end-to-end 
using standard backpropagation. Instead of evaluating each of the 16 gates 
individually (\textbf{FullEval} e.g., $A \cdot B$ for $A \wedge B$), we project this 
softmax-normalized gate distribution into a lower-dimensional functional 
basis space (\textbf{BasisProj}), enabling more efficient computation and improved interpretability. 
Specifically, the vector $\mathbf{p}$ is linearly projected via a fixed matrix 
$W_{16 \rightarrow 4} \in \mathbb{R}^{4 \times 16}$, yielding:
\begin{equation}
    \mathbf{c} = W_{16 \rightarrow 4} \cdot \mathbf{p}, \quad \text{where} \quad \mathbf{c} = [c_1, c_2, c_3, c_4]^\top
\end{equation}
are the coefficients for the basis $\{1, A, B, A \cdot B\}$.

The gate’s continuous output is then computed as:
\begin{equation}
    \text{output}_g = c_1 + c_2 A + c_3 B + c_4 (A \cdot B).
\end{equation}

The fixed projection matrix $W_{16 \rightarrow 4}$ is given by:

\begin{equation}
W_{16 \rightarrow 4}^\top =
\left[
\begin{array}{r|rrrr}
\text{Operator}           & 1 & A & B & A \cdot B \\
\hline
\text{False}              & 0 &  0 &  0 &  0 \\
A \land B                 & 0 &  0 &  0 &  1 \\
\lnot (A \Rightarrow B)   & 0 &  1 &  0 & -1 \\
A                         & 0 &  1 &  0 &  0 \\
\lnot (A \Leftarrow B)    & 0 &  0 &  1 & -1 \\
B                         & 0 &  0 &  1 &  0 \\
A \oplus B                & 0 &  1 &  1 & -2 \\
A \lor B                  & 0 &  1 &  1 & -1 \\
\lnot (A \lor B)          & 1 & -1 & -1 &  1 \\
\lnot (A \oplus B)        & 1 & -1 & -1 &  2 \\
\lnot B                   & 1 &  0 & -1 &  0 \\
A \Leftarrow B            & 1 &  0 & -1 &  1 \\
\lnot A                   & 1 & -1 &  0 &  0 \\
A \Rightarrow B           & 1 & -1 &  0 &  1 \\
\lnot (A \land B)         & 1 &  0 &  0 & -1 \\
\text{True}               & 1 &  0 &  0 &  0 \\
\end{array}
\right]
\label{eq:w_matrix_logical}
\end{equation}

This formulation offers several advantages. First, it allows efficient parallel computation using dense matrix operations, leveraging GPU acceleration. Implementation does not require platform specific (CUDA) code thus it is more portable. Second, it avoids evaluating 16 separate Boolean functions for each gate, improving training speed and reducing memory usage. Third, the decomposition into basis functions enhances interpretability by exposing the contribution of each term (e.g., $A$ vs.\ $A \cdot B$). Finally, it provides a flexible foundation for extending the logical expressivity to higher-order gates in future work.

\subsubsection{Learnable Interconnections}

For the learnable interconnect strategies (Top-$K$ and $L$), we similarly employ a softmax distribution over input weights to determine input pairings for each gate. For Top-$K$, a fixed number of candidates $K$ are preselected per gate at initialization, and only the softmax weights over those connections are trained. For the Fully Learnable strategy, the softmax is applied across all possible inputs, resulting in higher memory and computation costs.

% To control the softness of this selection process, we introduce a separate temperature parameter $\tau_c$ for connections. However, in our experiments we kept $\tau_c = 1$ and varied sparsity directly using the K value, which provides more explicit control. 

\subsubsection{Binarization}

After training, we convert the network into a discrete, inference-efficient form by binarizing both gate selection and interconnections. This is done by selecting the maximum probability option (the mode) from the softmax distributions. In practice, this means that each logic gate uses a fixed pair of binary inputs and computes a single Boolean function. Regardless of the training strategy, every node uses exactly two binary inputs at inference time, allowing the model to operate entirely in Boolean logic without any floating-point computation. 
% This discretization ensures that the model becomes extremely fast and memory-efficient during inference, benefiting applications in embedded or hardware-constrained environments.

\subsubsection{Classification Strategy}
We adopt a probabilistic majority-voting scheme for classification. The output layer consists of $n$ binary logic gates (e.g., 1,000), evenly divided into $k$ groups corresponding to the $k$ output classes (e.g., 10). During inference, each group counts the number of ‘1’ activations, and the predicted class is the one with the highest count. During training, soft outputs are aggregated per group, and the model is optimized using softmax cross-entropy loss. The sharpness of the soft logic activations is controlled by a global temperature parameter $\tau$, which was empirically tuned for each layer width. The parameter $\tau$ affects both convergence and final accuracy and was selected based on the results of a parameter sweep (see Sec.~\ref{sec:manualgain_experiments}). We use classification accuracy as the evaluation metric, as all tested datasets are balanced and accuracy effectively captures model performance.

%% file: sec/4_experiments.tex
\section{Experiments}

\subsection{Full Gate Evaluation vs.\ Basis Projection}

To assess the computational benefits of the proposed projection-based gate evaluation, we compare the \textbf{total training time} on the MNIST dataset (for an NVIDIA A4000 GPU) between two strategies:
\begin{itemize}
    \item \textbf{Full Gate Evaluation (FullEval):} Each of the 16 logic gates is evaluated individually per node.
    \item \textbf{Basis Projection (BasisProj):} A single projection into the \(\{1, A, B, A \cdot B\}\) basis using a fixed matrix \(W_{16 \rightarrow 4}\) replaces per-gate evaluation.
\end{itemize}

\begin{table}[ht!]
\centering
\caption{Full training time (in minutes) and speed-up using basis projection over full gate evaluation. Results are shown for the MNIST dataset and averaged over 10 runs with $\leq 0.1$ min variability.}
\begin{tabular}{cc|ccc}
\toprule
\textbf{Width} & \textbf{Type} & \textbf{FullEval} & \textbf{BasisProj} & \textbf{Speed-Up} \\
\midrule
  & 1F & 2.13  & 1.13 & 1.89$\times$ \\
8,000  & 2F & 4.75  & 1.60 & 2.96$\times$ \\
  & 3F & 7.47  & 2.21 & 3.38$\times$ \\
\midrule
 & 1F & 9.97 & 2.92 & 3.41$\times$ \\
32,000 & 2F & 20.78 & 5.44 & 3.82$\times$ \\
 & 3F & 31.63 & 7.99 & 3.96$\times$ \\
\bottomrule
\end{tabular}

\label{tab:training_time_full}
\end{table}

We evaluate six configurations, defined by the number of fixed connectome layers (1F, 2F, 3F) and layer width (8,000 and 32,000 nodes). As shown in Table~\ref{tab:training_time_full}, using basis projection significantly reduces total training time across all tested configurations. The benefit becomes increasingly pronounced with deeper and wider architectures. For instance, in the 32,000 / 3F configuration, the projection-based approach completes training nearly \textbf{4$\times$ faster} than the full gate evaluation. This speed-up stems from avoiding 16 separate logic gate computations per node in favor of a single dense matrix-vector multiplication—an operation highly optimized for modern accelerators. 
% Overall, these results demonstrate that our differentiable projection-based formulation is not only interpretable and expressive but also scalable and efficient for large models.

\subsection{Interconnect Strategy Analysis}

% We evaluate the performance of various interconnect strategies in LGNs across a broad spectrum of architectures, measuring binarized test-time accuracy as a function of layer width (2K–32K) and depth (1–4 layers). The test accuracy results are averaged over 10 runs for 1-2 layer models and 5 runs for 3-4 layer models. For completeness, we include all per-run statistics and training durations in the appendix.

We performed extensive experiments on the MNIST dataset using architectures with 1–4 layers and widths ranging from 2K to 32K, exploring various connectome variants. 
MNIST’s low computational cost allowed us to thoroughly analyze these designs, providing insights that informed the selection of a reduced set of architectures for FashionMNIST and CIFAR-10 experiments. 
Test accuracy was measured at binarized inference, averaged over at least 5 runs. 
% with per-run statistics and training times reported in the appendix. 
% All MNIST experiments were conducted on an NVIDIA A4000 GPU.

\subsubsection{Empirical Comparison Across Layer Widths and Depths on MNIST}

\begin{table}[ht]
  \centering
    \caption{Test accuracy results on MNIST across different layer widths and number of layers for various connectome methods. The results are averaged over at least 5 runs, all with $\leq 0.1\%$ variation. Bolded values indicate the best performance per block.}
  \begin{tabular}{@{}llcccc@{}}
    \toprule
    \makecell{\textbf{Layer} \\ \textbf{Width}} & \makecell{\textbf{Layer} \\ \textbf{Type}} & \textbf{1} & \textbf{2} & \textbf{3} & \textbf{4} \\
    \midrule
    \multirow{5}{*}{2,000} 
    & F       & 87.09 & 90.82 & 93.69 & 95.38 \\
    & Top8    & 94.29 & 96.97 & 97.47 & 97.59 \\
    & Top32   & 96.38 & 97.50 & \textbf{97.83} & \textbf{97.71} \\
    & Top128  & 97.02 & \textbf{97.66} &  97.67     &    96.92   \\
    & L       & \textbf{97.25} & 97.37 & 96.13 & 92.00 \\
    \midrule
    \multirow{5}{*}{4,000} 
    & F       & 90.16 & 93.28 & 95.77 & 96.90 \\
    & Top8    & 96.27 & 97.91 & 98.13 & 98.22 \\
    & Top32   & 97.54 & 98.17 & \textbf{98.38} & \textbf{98.29} \\
    & Top128  & 97.94 & \textbf{98.28} &    98.30   &   97.56    \\
    & L       & \textbf{97.96} & 98.11 & 97.02 &   94.41    \\
    \midrule
    \multirow{4}{*}{8,000} 
    & F       & 91.48 & 94.83 & 96.83 & 97.69 \\
    & Top8    & 97.17 & 98.43 & 98.60 & 98.63 \\
    & Top32   & 98.15 & 98.63 & \textbf{98.77} & \textbf{98.72} \\
    & Top128  & 98.39 & \textbf{98.67} &  98.72     &    97.94   \\
    & L       & \textbf{98.45} & 98.62 &  97.87 &  96.81     \\
    \midrule
    \multirow{5}{*}{16,000} 
    & F      & 93.16 & 96.43 & 97.82 & 98.34  \\
    & Top8   & 98.15 & 98.74 & 98.80 & \textbf{98.87} \\
    & Top32  & 98.50 & \textbf{98.95} & \textbf{98.86} & 98.73
  \\
    & Top128  & 98.55 & 98.89 &  98.84     &  98.23     \\
    & L       & \textbf{98.58} & 98.82 & - & -      \\
    \midrule
    \multirow{5}{*}{32,000} 
    & F       & 94.74 & 97.41 & 98.27 & 98.52  \\
    & Top8    & 98.41 & 98.75 & 98.78 & 98.76 \\
    & Top32   & \textbf{98.58} & 98.83 & \textbf{98.84} &  \textbf{98.89} \\
    & Top128  &   98.51 &   \textbf{98.91}   & 98.83 & 97.77      \\
    & L       & 98.52 & - & - &  -     \\
    \bottomrule
  \end{tabular}

  \label{tab:accuracy_layers}
\end{table}

\begin{figure}[h!]
    \centering
    \includegraphics[width=\linewidth]{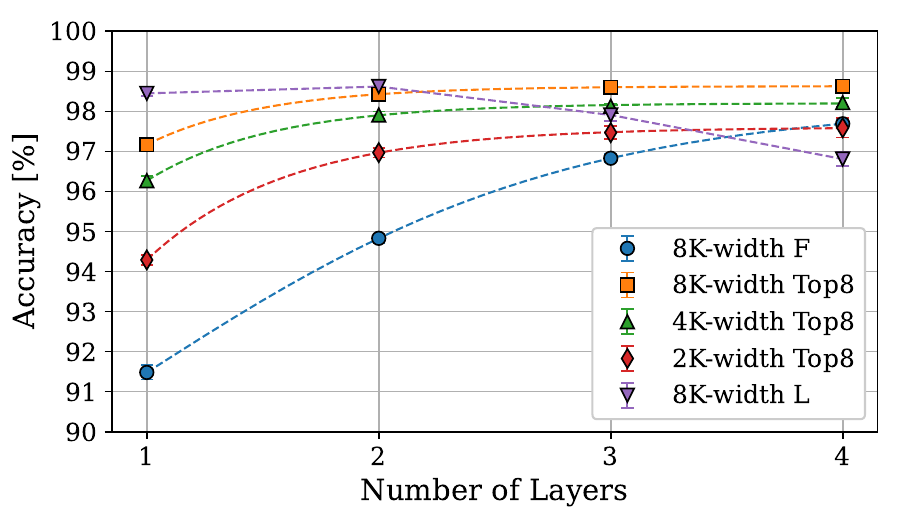}
    \caption{Test accuracy vs.\ depth for various interconnects.}
    \label{fig:plot_FKL}
\end{figure}

As shown in Table~\ref{tab:accuracy_layers}, random and non-trainable $F$ interconnects consistently perform the worst. Their inflexibility requires significantly wider or deeper networks to achieve moderate accuracy, illustrating the limitations of static wiring. Top-$K$ sparse interconnects, on the other hand, deliver strong results across nearly all configurations. Even under extreme sparsity (e.g., Top8), models reach over \textbf{98.6\%} accuracy when wide or deep enough, outperforming $L$ interconnects in many regimes. The best-performing configuration overall—a Top32 model with two 16K-wide layers (2Top32-16K)—achieves \textbf{98.95\%} accuracy. 
% Notably, Top-$K$ models continue to improve with additional depth, unlike $L$-type architectures, whose performance tends to saturate or decline beyond two layers due to overfitting or training instability.
Figure~\ref{fig:plot_FKL} illustrates how test accuracy scales with depth for different interconnect strategies and layer widths. 
% Notably, even under aggressive sparsity (Top8), models consistently outperform $F$ interconnects across all depths—even with reduced layer width. For instance, a 2Top8-2K model achieves 96.97\%, surpassing the 2F-8K model (94.83\%). $L$ interconnects attain slightly higher accuracy at shallow depths (e.g., 98.45\% for 1L-8K), but their gains saturate quickly. In contrast, Top8 models continue to improve with added depth, highlighting that sparse, learnable connectivity not only generalizes well but also scales more effectively than dense alternatives.

\begin{table*}[t!]
\centering
\caption{LILogicNet experimental results for different datasets and best performing architectures at given gate budgets.}
\label{tab:results}
\begin{tabular}{lrlc r r c r r}
\toprule
\textbf{Dataset} & \makecell{\textbf{Gate} \\ \textbf{Count}} & \textbf{Architecture} & \textbf{$\tau$} & \makecell{\textbf{Memory} \\ $$\textbf{[KB]}} & \makecell{\textbf{Train Time} \\ $$\textbf{[min]}} & \makecell{\textbf{Test Accuracy} \\ $$\textbf{[\%]}} & \makecell{\textbf{LUT-6s} \\ \textbf{(LGN)}} & \makecell{\textbf{LUT-6s (LGN} \\ \textbf{+popcount)}}\\
\midrule
\multirow{3}{*}{MNIST}
 & 4K   & 1L-4K        & 5  & 11.7 & 2.5 ± 0.0  & 97.96 ± 0.11 & 1,791 & 7,103 \\
 & 8K   & 1L-8K        & 10 & 23.4 & 4.3 ± 0.1 & 98.45 ± 0.07 & 3,491 & 14,076 \\
 & 32K  & 2Top32-16K   & 10 & 62.5 & 57.1 ± 0.2 & 98.95 ± 0.09 &  11,833 & 37,373 \\
\midrule
\multirow{3}{*}{FashionMNIST}
 & 8K   & 1L-8K        & 15  & 29.3 & 24.9 ± 0.2 &  89.95 ± 0.06  & 3,736 & 14,321 \\
 % & 32K  & 2Top32-16K   & 25  & 121.1 &  &  90.10 ± 0.07 \\
 & 64K  & 2Top32-32K   & 25  & 250.0 & 167.2 ± 2.7 &  90.26 ± 0.11 & 22,308 & 73,938  \\
 & 128K & 2Top32-64K   & 30  & 515.6 & 217.5 ± 3.4 & 90.61 ± 0.09 & 39,769 & 143,226\\
\midrule
\multirow{3}{*}{CIFAR10}
 & 8K   & 1L-8K        & 20  & 33.2 & 8.7 ± 0.1   & 55.11 ± 0.17 & 3,830 & 14,415 \\
 & 64K  & 1L-64K       & 90  & 266.0 & 62.9 ± 0.2  & 57.66 ± 0.17 & 27.070 & 104,853\\
 & 256K & 2Top32-128K  & 100 & 1,121.0 & 107.7 ± 0.1  & 60.98 ± 0.19 & 92,353 & 293,285\\
\bottomrule
\end{tabular}
\label{tab:lilogicnet_results}
\end{table*}

\subsubsection{Convergence of Top-K to Dense Connectivity}

\begin{figure}[ht!]
    \centering
    \includegraphics[width=\linewidth]{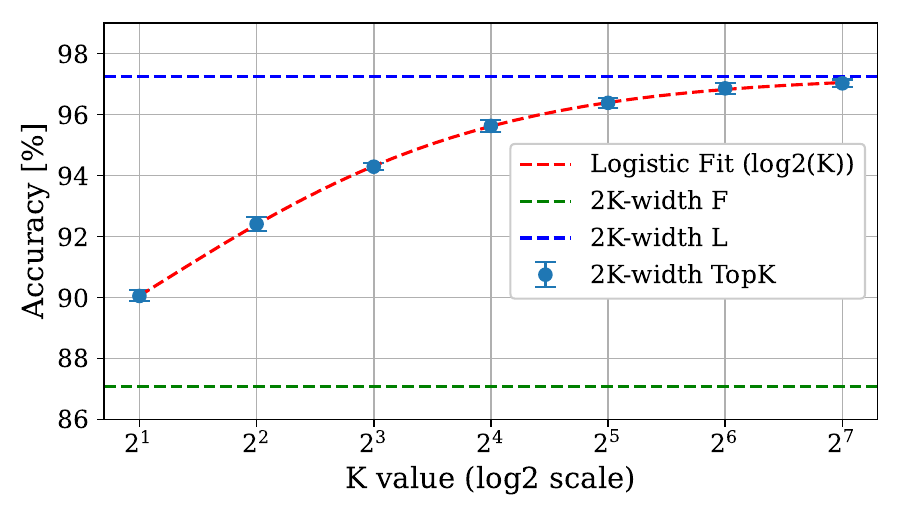}
    \caption{Convergence of 2K-width Top-$K$ test accuracy to $L$-type connectivity as $K$ increases.}
    \label{fig:ktop_convergence}
\end{figure}

Figure~\ref{fig:ktop_convergence} presents a representative case on the MNIST dataset of a 1-layer model with 2K gates, illustrating how Top-$K$ sparse interconnect performance improves with $K$, approaching that of the fully learnable (L) connectome. Accuracy rises quickly for small $K$ and saturates as $K$ increases; a logistic fit in $\log_2(K)$ space captures this trend. For $K \geq 128$, performance nearly reaches the fully learnable ceiling, reflecting the general trade-off between sparsity and accuracy.

% \subsubsection{Sparse Interconnects as Compositional Depth}
% Sparse interconnect strategies, such as Top-$K$, constrain each logic gate to operate on only a small subset of input pairs (e.g., 8 for Top-8). While this limits the immediate expressivity of a single layer, stacking multiple such layers expands the effective receptive field exponentially. Specifically, an $N$-layer network with Top-$K$ sparsity can, in theory, combine up to $K^N$ input paths — mimicking the connectivity of a fully learnable interconnect layer with $K^N$ input pairs. This binary tree–like composition enables sparse networks to approximate the flexibility of dense interconnects with much lower memory and compute requirements. For instance, two layers of Top-8 gates can model interactions over 64 input paths, equivalent to a single layer of Top-64 connectivity. Thus, adding depth to a sparse LGN serves as a powerful mechanism for capturing complex dependencies while maintaining hardware-efficient sparse connectivity.

\begin{figure}[h!]
\centering
\includegraphics[width=\linewidth]{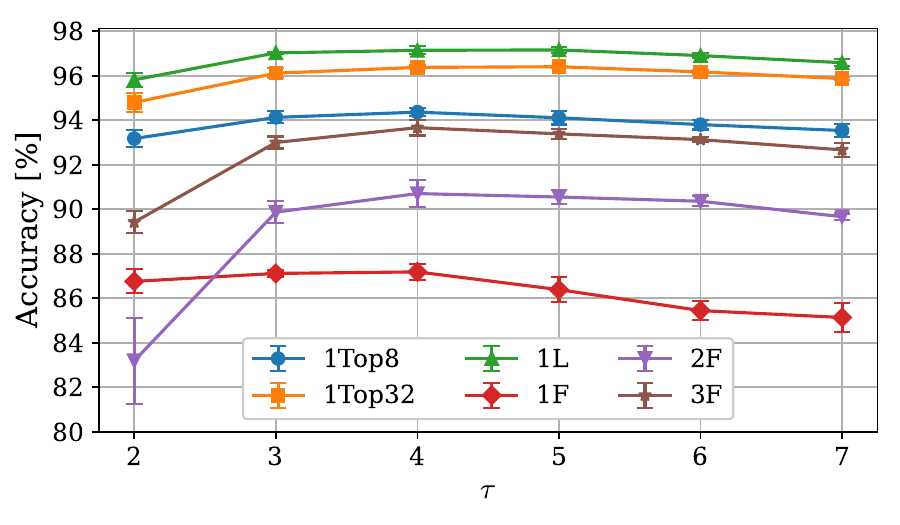}
\caption{Test accuracy as a function of $\tau$ for 2K-width models. Used as a reference for choosing temperature value.}
\label{fig:tau}
\end{figure}

\subsubsection{Sparse Interconnects as Compositional Depth}
Sparse interconnect strategies restrict each logic gate to operate on only a small subset of input pairs (e.g., 8 for Top8). While this limits the expressivity of a single layer, stacking layers expands the effective receptive field. In principle, an $N$-layer Top-$K$ network can access up to $K^N$ input paths---an upper bound that reflects the combinatorial breadth introduced by depth, even though each gate receives only two inputs at inference. Thus, depth enables sparse LGNs to recover some of the flexibility of denser interconnects at much lower cost. For example, two Top8 layers can, in the best case, reach up to 64 input paths, though the learned wiring may cover only part of this space. Depth therefore serves as an effective way to capture richer dependencies while maintaining hardware-efficient sparse connectivity.

\subsubsection{Sensitivity to Temperature Parameters}
\label{sec:manualgain_experiments}

We observe that the global softmax temperature 
$\tau$ should scale with layer width and dataset complexity (Table~\ref{tab:results}). For example, based on trends in Figure~\ref{fig:tau} for 2K-width MNIST architectures, we set $\tau = 4$. It is selected empirically to balance gradient flow during training and confident gate assignments after binarization: lower $\tau$ led to overly sharp distributions and vanishing gradients, while higher $\tau$ caused unstable training due to excessively flat distributions.

\begin{table}[t!]
\caption{Comparison of LGN models: FashionMNIST accuracy, FPGA resources, and latency (XC7A200T; XC7Z045 for DWN)}
\centering
\begin{tabular}{lcrr}
\toprule
\textbf{Model} & \textbf{Acc [\%]} & \makecell{\textbf{LUT-6s} \\ $$\textbf{[$\times 10^3$]}} & \makecell{\textbf{Latency} \\$$\textbf{[ns]}} \\
\midrule
DiffLogicNet-S* & 89.85 & 22.7 (10.4) & - \\
DiffLogicNet*   & 90.24 & 172.8 (72.8)  & - \\
DWN (n=6)     & 89.01 & \textbf{7.6} & 235 \\
LILogicNet-S    & 89.95 & 14.3 (3.7) & 30 \\
LILogicNet-M    & 90.26 & 73.9 (22.3) & 43 \\
LILogicNet-L    & \textbf{90.61} & 143.2 (39.8) & - \\
\bottomrule
\end{tabular}

\parbox{\linewidth}{\footnotesize * Results not reported in the original work; we reproduced the architecture using our framework with the same training setup and training time; \\
Values in parentheses correspond to the LGN part excluding the popcount; \\
- Could not fit on FPGA.
}
\label{tab:lgn_comparison}
\end{table}

\subsection{LILogic Net: Accuracy-Efficiency Pareto Front}

Based on accuracy, gate count, and training time, we identify three efficient design points for each dataset (Tab.~\ref{tab:lilogicnet_results}). These configurations provide strong predictive performance while maintaining low hardware cost and moderate training time, making them suitable for practical deployment. MNIST and FashionMNIST experiments were conducted on an NVIDIA A4000 GPU, while CIFAR-10 was trained on an NVIDIA H200 GPU to accommodate larger memory requirements for wider architectures. At smaller model scales, increasing layer width yields larger gains than increasing depth, while dense optimization utilizes limited capacity more effectively. However, as the gate budget increases, Top-$K$ models begin to outperform dense alternatives.

Reported training times include the full training procedure with validation but exclude initial data loading and online data augmentation. After enabling \texttt{torch.compile}, dense L-type models often train faster than sparse Top-$K$ variants, likely due to better GPU support for dense matrix operations compared to index-heavy sparse computations.

Parameter memory is reported assuming binarized LGNs at inference time with compact bit-packed storage of gate types and input indices. Each node stores its input connections and gate type, resulting in a memory cost proportional to the number of nodes and their fan-in.

% Based on accuracy, gate count, and training time, we identify three efficient design points for each dataset (Tab.~\ref{tab:lilogicnet_results}). These configurations provide strong predictive performance while maintaining low hardware cost and moderate training time, making them attractive candidates for practical deployment.

% The results indicate that at smaller model scales, increasing layer width tends to yield larger gains than increasing depth, while dense optimization helps utilize limited capacity more effectively. However, as the gate budget increases, Top-$K$ models begin to outperform dense alternatives.

% Reported training times include the complete training procedure with validation, but exclude the initial data loading and the online generation of augmented samples. 
% % This ensures that the reported values reflect only the forward and backward passes during training. 
% Interestingly, after enabling \texttt{torch.compile}, denser L-type models often train faster than their sparse Top-$K$ counterparts. Although fully learnable interconnects are theoretically more memory- and compute-intensive, this effect likely arises from the superior GPU support for dense matrix operations compared to sparse, index-heavy computations.

\label{tab:FPGA1}

% Training for MNIST and FashionMNIST was performed on an NVIDIA A4000 GPU, while CIFAR-10 experiments were conducted on an NVIDIA H200 GPU. The latter was chosen due to its larger memory capacity, which allowed us to evaluate wider architectures for the computationally expensive L-type layers. 
% Additionally, the CIFAR-10 input representation is substantially larger (21{,}504-dimensional), and keeping the dataset in GPU memory enabled more efficient experimentation.

% We report parameter memory assuming \textbf{binarized LGNs at inference time}, where gates and connections are stored in a compact bit-packed representation. Each node stores the indices of its input connections and the gate type. An input index requires $\lceil \log_2(L) \rceil$ bits, where $L$ is the number of possible inputs to the layer, while each gate requires 4 bits. The total parameter memory is therefore computed as the number of nodes multiplied by $\text{fan-in} \cdot \lceil \log_2(L) \rceil + \text{gate bits}$, summed across all layers.

We also report the number of LUT6 units required on Xilinx XC7A200T FPGA. We distinguish between the LUT usage for the LGN layers themselves and the total LUT usage including the final population count (popcount) stage. In practice, the popcount adders account for more than two-thirds of the total LUT consumption. To further improve FPGA efficiency, techniques such as the Learnable Reduction method proposed in DWN~\cite{bacellar2024differentiable} could be employed to significantly reduce this overhead.

\subsubsection{LILogicNet on FPGA}

% For comparable accuracy, our models require fewer LUT6 units than DiffLogic networks but more than DWN architectures, primarily due to the cost of the popcount stage (Tab.~\ref{tab:lgn_comparison}). Integrating the DWN Learnable Reduction approach could substantially decrease this overhead. Importantly, the proposed models were not optimized exclusively for FPGA deployment. For example, single-layer architectures are inherently inefficient in terms of LUT6 utilization, but they remain attractive when targeting a broader range of hardware platforms.

Table~\ref{tab:lgn_comparison} compares FPGA resource usage and latency of LGN models on FashionMNIST. For comparable accuracy, LILogicNet requires substantially fewer LUT6 units than DiffLogicNet, though more than DWN due to the cost of the popcount stage. Despite not being optimized specifically for FPGA deployment, LILogicNet achieves low inference latency, demonstrating its practical hardware efficiency.

 \subsubsection{LILogicNet on ASIC}
We taped out 6K-gate LILogicNet-S MNIST variant in SKY130 (SkyWater Technology 130 nm) and SG13 (IHP BiCMOS 130 nm)~\cite{lgn_mnist_repo,lgn_mnist_tinytapeout}, followed by an 8K-gate LILogicNet-S FashionMNIST variant in GF180 (GlobalFoundries 180 nm)~\cite{lgn_fashionmnist_repo, lgn_fashionmnist_tapeout}. The designs were implemented using the open-source OpenROAD~\cite{Ajayi2019OpenROADTA} flow with the respective open PDKs~\cite{sky130_pdk, sg13g2_pdk, gf180mcu_pdk}.

\subsubsection{Multi-Platform Hardware Throughput}

Table~\ref{tab:hardware_throughput} reports inference throughput on the GPU (NVIDIA A4000), CPU (Xeon Gold 6150, single thread), and FPGA (XC7A200T). The results show that LILogicNet maintains high inference throughput across diverse hardware platforms, reaching hundreds of thousands of FPS on GPU/CPU and tens of millions of FPS on FPGA for compact variants.

\begin{table}[ht!]
\caption{Inference throughput (t.) measured in frames per second (FPS) for FashionMNIST across different hardware platforms.}
\centering
\begin{tabular}{lrrr}
\toprule
Model & GPU t. & CPU t. & FPGA t. \\
\midrule
DiffLogicNet-S* & 145,472 & 812 & - \\
DiffLogicNet*   & 16,063 & 100 & - \\
LILogicNet-S    & 826,410 & 4,827 & 32,890,000 \\
LILogicNet-M    & 105,187 & 612 & 23,020,000 \\
LILogicNet-L    & 51,101 & 302 & - \\
\bottomrule
\end{tabular}
\label{tab:hardware_throughput}
\parbox{\linewidth}{\footnotesize * Results not reported in the original work; we reproduced the architecture using our framework with the same training setup and training time. \\
- Could not fit on FPGA.}
\end{table}

% \subsection{Comparison with previous work}

% Table~\ref{tab:accuracy_vs_gates} compares LILogicNet with prior logic-based models on the MNIST and CIFAR-10 datasets.
% For MNIST, our smallest model, LILogicNet-S, achieves 97.96\% accuracy using just 4K gates—outperforming DiffLogic Net-S, which requires 48K gates for 97.69\%. LILogicNet-M reaches 98.45\% with 8K gates, matching substantially larger models such as LogicTreeNet-S (147K) and DiffLogic Net (384K). LILogicNet-L achieves 98.95\% accuracy with only 32K gates, approaching the performance of LogicTreeNet-M and -L, which use 566K and 1.27M gates, respectively. While RV-LGN and eXpLogic also operate at low gate counts (18K and 5K), they suffer from significantly lower accuracy (95.80\% and 91.20\%), highlighting the challenge of maintaining performance in highly compact designs. In contrast, LILogicNet models extend the Pareto frontier by offering competitive accuracy at minimal hardware cost.

% For CIFAR-10, all LILogicNet variants—S, M, and L—outperform their corresponding DiffLogic Net baselines (S, M, and L) while using an order of magnitude fewer gates. The largest variant, LILogicNet-L, also surpasses LogicTreeNet-S in accuracy, despite being much smaller. Although the largest DiffLogic Net and LogicTreeNet models (LogicTreeNet-B and -L) achieve higher accuracy, they rely on much larger architectures. In this work, we limited our exploration to models of up to 256K gates due to hardware memory constraints, which we plan to address in future extensions to evaluate larger configurations.

\begin{table}[h]
\caption{Accuracy vs. gate count and FPGA throughput comparison for LGN models on MNIST, CIFAR-10, and FashionMNIST.}
\centering
\begin{tabular}{llrr}
\toprule

\textbf{Model}  & \textbf{Acc [\%]} & \makecell{\textbf{Gate} \\ \textbf{Count}} & \makecell{\textbf{FPGA t.} \\ $$\textbf{[FPS]}} \\
\midrule
\multicolumn{4}{c}{\textbf{MNIST}} \\
\midrule
LogicTreeNet-S~\cite{petersen2024convolutional} & 98.46 & 147 K & \textbf{250.0 M} \\
LogicTreeNet-M~\cite{petersen2024convolutional} & 99.23 & 566 K & \textbf{200.0 M} \\
LogicTreeNet-L~\cite{petersen2024convolutional} & \textbf{99.35} & 1.27 M & -\\
\midrule
DiffLogicNet-S~\cite{petersen2022deep} & 97.69 & 48 K & -\\
DiffLogicNet~\cite{petersen2022deep} & 98.47 & 384 K & -\\
RV-LGN~\cite{wang2025logic} & 95.80 & 18 K & - \\
eXpLogic~\cite{wormald2025explogic} & 91.20 & 5 K & - \\
DWN(n=6)~\cite{bacellar2024differentiable} & 98.77 & - & 22.2 M\\
\textbf{LILogicNet-S (ours)} & 97.96 & \textbf{4 K} & \textbf{40.1 M} \\
\textbf{LILogicNet-M (ours)} & 98.45 & 8 K & 32.9 M \\
\textbf{LILogicNet-L (ours)} & \textbf{98.95} & 32 K & 29.7 M \\
\midrule
\multicolumn{4}{c}{\textbf{FashionMNIST}} \\
\midrule
% \textbf{Method} & \textbf{Acc [\%]} & \textbf{Gates Count} & \makecell{\textbf{FPGA t.} \\ [samples/s]} \\
% \midrule
DiffLogicNet-S*~\cite{petersen2022deep} & 89.85 & 48 K & -\\
DiffLogicNet*~\cite{petersen2022deep} & 90.24 & 384 K & -\\
DWN(n=2)~\cite{bacellar2024differentiable} & 89.12 & - & 10.0 M\\
DWN(n=6)~\cite{bacellar2024differentiable} & 89.01 & - & 10.0 M\\
\textbf{LILogicNet-S (ours)} & 89.95 & \textbf{8 K} & \textbf{32.9 M} \\
\textbf{LILogicNet-M (ours)} & 90.26 & 64 K & 23.0 M\\
\textbf{LILogicNet-L (ours)} & \textbf{90.61} & 128 K & -\\
\midrule
\multicolumn{4}{c}{\textbf{CIFAR-10}} \\
\midrule
% \textbf{Method} & \textbf{Acc [\%]} & \textbf{\#Gates} & \makecell{\textbf{$\Delta$Acc/G} \\ $[\times 10^{-5}]$} \\
% \midrule
% FINN CNV~\cite{umuroglu2017finn} & 80.10 & 901 M & 0.003 \\
LogicTreeNet-S~\cite{petersen2024convolutional} & 60.38 & 400 K & \textbf{111.1 M} \\
LogicTreeNet-G~\cite{petersen2024convolutional} & 86.21 & 61 M & - \\
\midrule
DiffLogicNet-S~\cite{petersen2022deep} & 51.27 & 48 K & -\\
DiffLogicNet-M~\cite{petersen2022deep} & 57.39 & 512 K & -\\
DiffLogicNet-L~\cite{petersen2022deep} & 60.78 & 1.28 M & - \\
DiffLogicNet-XL~\cite{petersen2022deep} & 62.14 & 5.12 M & - \\

% TTNet-S~\cite{benamira2024truth} & 50.10 & 565 K & 0.02 \\
% TTNet-L~\cite{benamira2024truth} & 70.75 & 189 M & 0.01 \\
RV-LGN~\cite{wang2025logic} & 51.30 & 36 K & - \\
DWN(n=2)~\cite{bacellar2024differentiable} & 57.51 & - & 468 K \\
DWN(n=6)~\cite{bacellar2024differentiable} & 57.42 & - & 468 K\\
\textbf{LILogicNet-S (ours)} & 55.11 & \textbf{8 K} & \textbf{31.2 M}  \\
\textbf{LILogicNet-M (ours)} & 57.66 & 64 K & - \\
\textbf{LILogicNet-L (ours)} & \textbf{60.98} & 256 K & - \\

\bottomrule
\end{tabular}
\parbox{\linewidth}{\footnotesize * Results not reported in the original work; we reproduced the architecture using our framework and the same training setup and training time. \\
- Value not reported or could not fit on FPGA.}
\\
\parbox{\linewidth}{\footnotesize }
\label{tab:accuracy_vs_gates}
\end{table}

\subsection{Comparison with previous work}

Table~\ref{tab:accuracy_vs_gates} compares LILogicNet with prior logic-based models in terms of accuracy, gate count, and FPGA throughput across MNIST, FashionMNIST, and CIFAR-10. Overall, LILogicNet achieves competitive accuracy while requiring substantially fewer gates than prior LGN architectures. For example, on MNIST, LILogicNet-M achieves 98.45\% accuracy with only 8K gates, compared to 384K gates for DiffLogicNet with similar accuracy. Similar trends hold on FashionMNIST, where LILogicNet slightly improves accuracy over DiffLogicNet while using significantly smaller models. On CIFAR-10, LILogicNet consistently outperforms DiffLogicNet variants under comparable gate budgets. In addition, LILogicNet achieves higher FPGA throughput than DWN despite being evaluated on a smaller FPGA and without hardware-specific optimization, indicating strong potential for efficient deployment.

LogicTreeNet differs from the remaining approaches as it adopts a CNN-like architecture, while the rest LGN models follow a feed-forward design. As a result, its larger variants achieve substantially higher accuracy but require significantly larger gate budgets. In this work, we limited our exploration to models up to 256K gates, where LILogicNet achieves competitive or better accuracy than the smallest LogicTreeNet models with fewer gates. Scaling learnable connectomes to larger architectures remains challenging due to the combinatorial growth of possible logic configurations, and will be investigated in future work.

% Potential improvements
% Higher accuracy is likely achievable with longer training, kernel-level optimizations, and additional regularization.

%% file: sec/5_conclusion.tex
\section{Conclusion}

% We introduced \textbf{LILogic Net}, an efficient logic gate network that jointly learns gate functions and their interconnections. By optimizing the connectome during training and employing a projection-based gate evaluation, our models achieve fast, stable convergence without compromising accuracy. For instance, the 8K-gate LILogicNet-M reaches 98.45\% on MNIST in only 4.3 minutes, outperforming prior logic-based models that use up to two orders of magnitude more gates.
% Our design favors simplicity over heavy regularization or handcrafted initialization, yet the models generalize strongly, indicating substantial untapped optimization potential. Future extensions—such as dropout, weight decay, alternative scaling, or custom normalization—could further enhance robustness and training efficiency. 
% Although CIFAR-10 experiments were limited by hardware, they already demonstrate scalability to more complex data. Planned improvements with custom-optimized kernels will enable larger models and deeper architectures.
% Overall, LILogic Net lays a strong foundation for efficient, interpretable, and hardware-friendly logic networks, bridging symbolic computation and modern neural design.

We introduced \textbf{LILogicNet}, a compact logic gate network that jointly learns gate functions and their interconnections. Top-$K$ connectivity enables the model to focus on the most informative connections, improving accuracy and efficiency compared to fixed or dense designs. Basis Projection further speeds up training by up to 4× by efficiently computing gate outputs via a fixed projection. LILogicNet maintains high inference throughput across GPU, CPU, and FPGA, with compact variants delivering hundreds of thousands to tens of millions of FPS, and its performance could be further improved by hardware-specific optimizations. Future work will explore regularization and architectural reductions (e.g., gates or adders) to scale to larger models and more complex datasets. Overall, LILogicNet provides an efficient, interpretable, and hardware-friendly framework for learnable logic networks, with potential for further improvements in performance, efficiency, and scalability.

\paragraph{Acknowledgments} This work was supported by the Department of Artificial Intelligence at the Wrocław University of Science and Technology.

%% file: sec/X_suppl.tex
\clearpage
\setcounter{page}{1}
\maketitlesupplementary

\begin{table*}[th!]
  \centering
  \small
  \setlength{\tabcolsep}{5.5pt}
  \begin{tabular}{@{}llcccccccc@{}}
    \toprule
    \textbf{Width} & \textbf{Type} & \multicolumn{2}{c}{\textbf{1 Layer}} & \multicolumn{2}{c}{\textbf{2 Layers}} & \multicolumn{2}{c}{\textbf{3 Layers}} & \multicolumn{2}{c}{\textbf{4 Layers}} \\
    \cmidrule(lr){3-4} \cmidrule(lr){5-6} \cmidrule(lr){7-8} \cmidrule(lr){9-10}
    & & \makecell{Test \\ Acc. [\%]} & \makecell{Train \\Time [min]} & \makecell{Test \\ Acc. [\%]} & \makecell{Train \\Time [min]} & \makecell{Test \\ Acc. [\%]} & \makecell{Train \\Time [min]} & \makecell{Test \\ Acc. [\%]} & \makecell{Train \\Time [min]} \\
    \midrule
    \multirow{9}{*}{2,000}
    & F        & 87.09 ± 0.28 & 3.3 ± 0.3 & 90.82 ± 0.40 & 5.5 ± 0.6 & 93.69 ± 0.29 & 7.4 ± 0.4 & 95.38 ± 0.20 & 8.9 ± 0.7\\
    & Top2     & 90.05 ± 0.18 & 4.3 ± 0.2 & 94.58 ± 0.21 & 1.9 ± 0.0  & 96.25 ± 0.12  & 2.5 ± 0.0 & 96.73 ± 0.12  & 3.1 ± 0.0  \\
    & Top4     & 92.41 ± 0.23 & 4.6 ± 0.3 & 96.39 ± 0.15 & 2.0 ± 0.0  & 97.19 ± 0.12  & 2.7 ± 0.1  & 97.38 ± 0.14  & 3.3 ± 0.1  \\
    & Top8     & 94.29 ± 0.12 & 4.4 ± 0.3 & 96.97 ± 0.12 & 2.0 ± 0.0 & 97.47 ± 0.16 & 2.7 ± 0.0 & 97.59 ± 0.24 & 3.4 ± 0.0 \\
    & Top16    & 95.63 ± 0.19  & 4.6 ± 0.3  & 97.35 ± 0.08  & 2.6 ± 0.0 & 97.72 ± 0.12 & 4.0 ± 0.0 & \textbf{97.75 ± 0.12} & 5.3 ± 0.0 \\
    & Top32    & 96.38 ± 0.15 & 4.6 ± 0.2 & 97.50 ± 0.13 & 4.8 ± 0.0 & 97.83 ± 0.15 & 7.8 ± 0.0 & 97.71 ± 0.12 & 10.8 ± 0.0 \\
    & Top64    & 96.86 ± 0.18  & 5.1 ± 0.3 & 97.65 ± 0.05 & 8.9 ± 0.0 & \textbf{97.89 ± 0.06} & 15.1 ± 0.0 & 97.45 ± 0.31 & 21.3 ± 0.0 \\
    & Top128   & 97.02 ± 0.12 & 5.6 ± 0.1 & \textbf{97.66 ± 0.06} & 17.6 ± 0.0 & 97.67 ± 0.17 & 23.3 ± 0.0 & 96.92 ± 0.14 & 37.5 ± 6.5 \\
    & L        & \textbf{97.25 ± 0.13} & 5.0 ± 0.4 & 97.37 ± 0.08 & 4.1 ± 0.0 & 96.13 ± 0.52 & 6.4 ± 0.0 & 92.00 ± 1.54 & 8.5 ± 0.1 \\
    \midrule
    \multirow{9}{*}{4,000}
    & F        & 90.16 ± 0.24 & 1.4 ± 0.6 & 93.28 ± 0.21 & 1.9 ± 0.7 & 95.77 ± 0.18 & 2.4 ± 1.0 & 96.90 ± 0.11 & 2.7 ± 0.0 \\
    & Top2     & 92.37 ± 0.14 & 1.5 ± 0.4 & 96.50 ± 0.15 & 2.0 ± 0.1 & 97.56 ± 0.13 & 2.6 ± 0.1 & 97.84 ± 0.15 & 3.2 ± 0.0 \\
    & Top4     & 94.66 ± 0.26 & 1.5 ± 0.0 & 97.61 ± 0.10 & 2.2 ± 0.1 & 97.99 ± 0.10 & 2.9 ± 0.0 & 98.04 ± 0.21 & 3.7 ± 0.0 \\
    & Top8     & 96.27 ± 0.11 & 1.5 ± 0.0 & 97.91 ± 0.09 & 2.8 ± 0.0 & 98.13 ± 0.07 & 4.2 ± 0.0 & 98.22 ± 0.11 & 5.7 ± 0.0 \\
    & Top16    & 97.17 ± 0.10 & 1.9 ± 0.0 & 98.14 ± 0.15 & 4.6 ± 0.1 & 98.33 ± 0.09 & 7.3 ± 0.0 & \textbf{98.37 ± 0.09} & 10.1 ± 0.0 \\
    & Top32    & 97.54 ± 0.16 & 3.1 ± 0.9 & 98.17 ± 0.16 & 8.3 ± 0.1 & \textbf{98.38 ± 0.05} & 13.8 ± 0.0 & 98.29 ± 0.12 & 19.4 ± 0.1 \\
    & Top64    & 97.76 ± 0.09 & 6.3 ± 0.0 & 98.25 ± 0.07 & 19.0 ± 0.0 & 98.31 ± 0.09 & 31.5 ± 0.1 & 98.06 ± 0.18 & 44.1 ± 0.0 \\
    & Top128   & 97.94 ± 0.07 & 8.0 ± 0.0 & \textbf{98.28 ± 0.08} & 33.1 ± 0.2 & 98.30 ± 0.07 & 58.8 ± 0.0 & 97.56 ± 0.16 & 83.6 ± 0.1 \\
    & L        & \textbf{97.96 ± 0.11} & 2.5 ± 0.0 & 98.11 ± 0.06 & 10.8 ± 0.0 & 97.02 ± 0.53 & 19.2 ± 0.1 & 94.41 ± 1.49 & 27.4 ± 0.3 \\
    \midrule
    \multirow{9}{*}{8,000}
    & F        & 91.48 ± 0.18 & 1.1 ± 0.0 & 94.83 ± 0.12 & 1.6 ± 0.0 & 96.83 ± 0.11 & 2.2 ± 0.0 & 97.69 ± 0.12 & 2.8 ± 0.0 \\
    & Top2     & 93.72 ± 0.09 & 1.2 ± 0.0 & 97.41 ± 0.07 & 2.7 ± 0.0 & 98.20 ± 0.11 & 4.0 ± 0.0 & 98.28 ± 0.14 & 5.3 ± 0.0 \\
    & Top4     & 95.70 ± 0.12 & 1.4 ± 0.0  & 98.24 ± 0.08 & 3.4 ± 0.0 & 98.50 ± 0.04 & 5.4 ± 0.0 & 98.56 ± 0.05 & 7.3 ± 0.0 \\
    & Top8     & 97.17 ± 0.07 & 1.8 ± 0.0 & 98.43 ± 0.07 & 5.7 ± 0.0 & 98.60 ± 0.06 & 9.4 ± 0.0 & 98.63 ± 0.09 & 13.1 ± 0.0 \\
    & Top16    & 97.85 ± 0.08 & 2.8 ± 0.0  & 98.66 ± 0.06 & 10.3 ± 0.0 & 98.69 ± 0.04 & 17.6 ± 0.1 & \textbf{98.79 ± 0.07} & 24.9 ± 0.1 \\
    & Top32    & 98.15 ± 0.02 & 6.3 ± 0.0 & 98.63 ± 0.08 & 22.8 ± 0.1 & \textbf{98.77 ± 0.13} & 39.2 ± 0.1 & 98.72 ± 0.05 & 55.5 ± 0.2 \\
    & Top64    & 98.36 ± 0.07 & 11.6 ± 0.0  & \textbf{98.69 ± 0.06} & 45.3 ± 0.2 & 98.74 ± 0.09 & 78.5 ± 0.3 & 98.44 ± 0.10 & 111.5 ± 0.4 \\
    & Top128   & 98.39 ± 0.07 & 16.2 ± 0.1 & 98.67 ± 0.06 & 82.4 ± 0.4 & 98.72 ± 0.11 & 148.1 ± 0.3 & 97.94 ± 0.28 & 213.8 ± 0.6 \\
    & L        & \textbf{98.45 ± 0.07} & 4.3 ± 0.1 & 98.62 ± 0.12 & 4.4 ± 0.1 & 97.87 ± 0.22 & 69.2 ± 0.2 & 96.81 ± 0.77 & 101.4 ± 0.2 \\
    \midrule
    \multirow{9}{*}{16,000}
    & F        & 93.16 ± 0.16 & 1.3 ± 0.0 & 96.43 ± 0.12 & 2.6 ± 0.0 & 97.82 ± 0.13 & 3.8 ± 0.0 & 98.34 ± 0.08 & 5.1 ± 0.0 \\
    & Top2     & 95.52 ± 0.16 & 1.8 ± 0.0  & 98.15 ± 0.09 & 6.0 ± 0.1 & 98.53 ± 0.03 & 10.0 ± 0.0 & 98.64 ± 0.09 & 14.0 ± 0.0 \\
    & Top4     & 97.24 ± 0.09 & 2.2 ± 0.0  & 98.59 ± 0.08 & 8.8 ± 0.0 & 98.68 ± 0.06 & 15.3 ± 0.0 & 98.77 ± 0.06 & 21.8 ± 0.1 \\
    & Top8     & 98.15 ± 0.04 & 3.1 ± 0.0 & 98.74 ± 0.10 & 14.7 ± 0.1 & 98.80 ± 0.08 & 25.9 ± 0.1 & \textbf{98.87} ± 0.08 & 37.3 ± 0.4 \\
    & Top16    & 98.48 ± 0.10 & 6.9 ± 0.0  & 98.80 ± 0.05 & 30.2 ± 0.1 & 98.79 ± 0.08 & 53.4 ± 0.2 & 98.83 ± 0.10 & 76.7 ± 0.5 \\
    & Top32    & 98.50 ± 0.07 & 12.2 ± 0.0 & \textbf{98.95 ± 0.09} & 57.1 ± 0.2 & \textbf{98.86 ± 0.07} & 101.6 ± 0.3 & 98.73 ± 0.05 & 146.2 ± 0.5 \\
    & Top64    & 98.53 ± 0.11 & 22.5 ± 0.0  & 98.88 ± 0.06 & 111.9 ± 0.3 & 98.85 ± 0.10 & 200.7 ± 0.4 & 98.71 ± 0.04 & 289.1 ± 0.3 \\
    & Top128   & 98.55 ± 0.03 & 42.3 ± 0.0 & 98.89 ± 0.11 & 219.4 ± 0.3 & 98.84 ± 0.09 & 396.5 ± 0.3 & 98.23 ± 0.12 & 475.6 ± 0.5 \\
    & L        & \textbf{98.58} ± 0.07 & 7.8 ± 0.0 & 98.82 ± 0.10 & 154.9 ± 0.1 & --           & --          & --           & --          \\
    \midrule
    \multirow{9}{*}{32,000}
    & F        & 94.74 ± 0.07 & 2.9 ± 0.0 & 97.41 ± 0.13 & 5.4 ± 0.0 & 98.27 ± 0.10 & 8.0 ± 0.1 & 98.52 ± 0.09 & 10.4 ± 0.2 \\
    & Top2     & 96.85 ± 0.06 & 3.1 ± 0.0  & 98.41 ± 0.12 & 15.8 ± 0.1 & 98.64 ± 0.11 & 28.7 ± 0.2 & 98.58 ± 0.06 & 41.0 ± 0.7 \\
    & Top4     & 98.05 ± 0.11 & 4.0 ± 0.0  & 98.67 ± 0.05 & 23.0 ± 0.1 & 98.81 ± 0.04 & 43.6 ± 0.3 & 98.79 ± 0.08 & 61.4 ± 0.6 \\
    & Top8     & 98.41 ± 0.11 & 7.7 ± 0.0 & 98.75 ± 0.07 & 40.5 ± 0.1 & 98.78 ± 0.01 & 69.1 ± 0.3 & 98.76 ± 0.06 & 105.5 ± 0.3 \\
    & Top16    & 98.56 ± 0.07 & 13.4 ± 0.1  & 98.83 ± 0.04 & 57.9 ± 1.0 & 98.83 ± 0.05 & 103.0 ± 0.5 & 98.81 ± 0.05 & 149.5 ± 0.5 \\
    & Top32    & \textbf{98.58 ± 0.07} & 24.2 ± 0.1 & 98.83 ± 0.09 & 134.3 ± 0.1 & 98.84 ± 0.10 & 240.7 ± 1.2 & \textbf{98.89 ± 0.06} & 348.4 ± 0.7 \\
    & Top64    & 98.55 ± 0.10 & 45.5 ± 0.1  & 98.87 ± 0.07 & 268.0 ± 0.8 & \textbf{98.87 ± 0.05} & 490.2 ± 6.6 & 97.04 ± 0.04 & 723.0 ± 1.2 \\
    & Top128   & 98.51 ± 0.08 & 84.0 ± 0.1 & \textbf{98.91 ± 0.05} & 524.1 ± 0.2 & 98.83 ± 0.04 & 884.5 ± 4.2 & 97.77 ± 0.20 & 1326.6 ± 5.1 \\
    & L        & 98.52 ± 0.09 & 15.0 ± 0.1 & --           & --          & --           & --          & --           & --          \\
    \bottomrule
  \end{tabular}
  \caption{Test accuracy and training time on MNIST for different logic gate networks, layer widths, and depths. Each cell shows the mean ± standard deviation over multiple runs (10 runs for 1-2 layers, 5 runs for 3-4 layers). Dashes indicate inapplicable configurations. Bolded values indicate the best performance per block.  \newline }
  \label{tab:accuracy_layers_full}
\end{table*}

\newpage

\section{Full Experimental Results on MNIST}
\label{sec:mnist_full_results}

Table~\ref{tab:accuracy_layers_full} presents the complete set of results from our experiments on the MNIST dataset, expanding on the summary shown in the main paper.

We systematically varied the \textbf{layer width}, \textbf{network depth} (1 to 4 layers), and \textbf{logical connectivity strategy}, including:
\begin{itemize}
    \item \textbf{F}: Fixed, non-learnable connectome,
    \item \textbf{Top-K}: Top-$K$ sparse, learnable connectome with $K$ $\in$ $\{2, 4, 8, 16,$ $32, 64, 128\}$,
    \item \textbf{L}: Dense, fully learnable connectome.
\end{itemize}

\noindent For each architecture, we report:
\begin{itemize}
    \item \textbf{Test accuracy} (mean $\pm$ std.\ deviation - in \%),
    \item \textbf{Training time} (mean $\pm$ std.\ deviation - in minutes).
\end{itemize}

\subsection{Experimental Setup}
\begin{itemize}
    \item Each configuration was run 10 times for 1- and 2-layer models, and 5 times for 3- and 4-layer networks.
    \item Training time was measured on a single NVIDIA A4000 GPU and includes model validation every 25 epochs (200 in total).
    \item Entries with ``--'' indicate configurations that were infeasible due to memory constraints.
\end{itemize}

\subsection{Key Observations}
% \begin{itemize}

\paragraph{Learned sparse Top-$K$ connectivity consistently outperforms fixed random connectivity, especially at smaller model widths, with the performance gap narrowing as the width increases}  
At lower widths (e.g., 2K), the accuracy difference between Top-16 and fixed random (F) connectivity is more pronounced (over 2\% points difference at 4 layers). As width increases to 32K, both methods achieve high accuracy, with Top-$K$ maintaining a modest advantage (around 0.3–0.4\% point). This indicates that while fixed random wiring can suffice at large scales, learned sparse wiring helps especially in smaller or medium-width models by optimizing logical connections more effectively.

\paragraph{Sparse Top-$K$ connectivity improves accuracy across widths, with depth effects depending on $K$}  
Increasing the number of inputs $K$ per gate generally improves accuracy across all widths (2K to 32K). For smaller $K$ values (e.g., Top-2, Top-4), increasing the number of layers consistently enhances performance, reflecting the benefit of deeper feature integration. However, for larger $K$ values (e.g., Top-64, Top-128), the best accuracy is often achieved at 2 or 3 layers, with 4-layer models showing little or no further improvement—or even slight degradation. This suggests that when gates have access to more inputs, very deep architectures may lead to overfitting or diminishing returns in this setting.

\paragraph{Top-$K$ offers an efficient trade-off between accuracy and computational cost}  
Top-$K$ connectivity sparsifies each gate’s inputs, significantly reducing the number of active computations during both training and inference, while retaining high accuracy. For instance, at 8K width and 2 layers, a Top-16 model achieves \textbf{98.66\%} accuracy — slightly exceeding fully learnable (L) models — while using fewer connections and benefiting from reduced computational complexity. This makes Top-$K$ particularly attractive for efficient, scalable architectures in resource-constrained settings.

\paragraph{Comparison of fully learned (L) and Top-$K$ connectivity within the same width and depth}  
At shallow depths (e.g., 1 layer), learned connectivity (L) consistently achieves the highest accuracy, outperforming both fixed random (F) and Top-$K$ sparsity patterns. However, as the network depth increases, Top-$K$ connectivity increasingly surpasses learned connectivity, demonstrating stronger scalability and better generalization in deeper models. Interestingly, for deeper architectures, smaller values of $K$ in Top-$K$ (e.g., Top-8, Top-16) tend to yield the best accuracy, suggesting that moderate sparse wiring is sufficient for effective information integration while controlling complexity and overfitting.

\paragraph{At fixed gate budgets, shallow and wide architectures often outperform deeper ones — until a certain scale is reached}  
As discussed in Section 4.3, under constrained gate budgets (2K–8K), shallow fully learnable ($L$) models achieve surprisingly strong performance. For instance, a 1-layer 4K-width model (1L–4K) reaches \textbf{97.96\%} accuracy, outperforming deeper or sparse Top-$K$ models of similar budget (e.g., 2Top128–2K with \textbf{97.66\%}). This suggests that when capacity is limited, wider layers and dense training provide more effective use of resources than depth or sparsity. However, at higher budgets (e.g., 8K+), Top-$K$ architectures begin to outperform learned ones. For example, 2Top64–8K reaches \textbf{98.69\%}, exceeding 1L–16K (\textbf{98.58\%}). Depth becomes beneficial beyond a threshold, but results show that 2-layer configurations often outperform their 4-layer counterparts (e.g., 2Top64–8K vs.\ 4Top16–4K), indicating diminishing returns from excessive depth once sparsity and width are well-balanced.

\paragraph{Training time scales non-uniformly with width and $K$, but grows roughly linearly with depth}
Training time increases with model width, but the scaling is non-monotonic: from 2K to 8K gates, the time increase is modest, likely due to better GPU utilization. However, for larger widths (16K–32K), the increase becomes steeper, suggesting that memory or computational bottlenecks begin to dominate.
The relationship between training time and the number of gate inputs $K$ is not strictly linear: while larger $K$ generally increases cost, the overhead is influenced by implementation details, such as sparse gather operations and input aggregation. Top-128 models are much slower than Top-16, but the scaling trend is irregular.
In contrast, training time grows approximately linearly with depth — adding layers increases computation in a predictable way.
Notably, fully learnable connectivity (L) often trains faster than Top-$K$ models despite involving weight optimization. This is likely due to GPU-accelerated dense matrix operations (\texttt{matmul}) being more efficient than sparse input selection used in Top-$K$.
Several anomalies are observed:
(i) For width 2K, 1-layer model, training times are longer than for 4K or 8K models. This may stem from poor GPU utilization, kernel launch overhead, or small matrix sizes being inefficiently handled.
(ii) Fixed connectivity (F) also shows unexpectedly long training times at low widths (e.g., 2K), possibly due to inefficient memory access or unoptimized computation of fixed connections.

\paragraph{Low variance in results confirms robustness}  
Across all settings, standard deviation in test accuracy remains under 0.2\%, indicating stable convergence and reproducibility, even for models with high structural sparsity.

\section{Full Experimental Results on CIFAR-10}
\label{sec:cifar_full_results}

Table~\ref{tab:cifar_combined} presents the complete set of results from our experiments on the CIFAR-10 dataset, expanding on the summary shown in the main paper.

Building on insights from MNIST, we focused on 1-layer variants for smaller architectures (8K gates) and 1–2-layer variants for wider layers. We evaluated the following logical connectivity strategies:
\begin{itemize}
\item \textbf{F}: Fixed, non-learnable connectome,
\item \textbf{Top-K}: Top-$K$ sparse, learnable connectome with $K=32$, chosen based on MNIST experiments as a good trade-off between accuracy and training time,
\item \textbf{L}: Dense, fully learnable connectome.
\end{itemize}

We also experimented with two sets of fixed thresholds to binarize each RGB channel:
\begin{itemize}
\item $[0.125, 0.25, 0.375, 0.5, 0.625, 0.75, 0.875]$,
\item $[0.2, 0.4, 0.6, 0.8]$,
\end{itemize}
resulting in input vector sizes of 21,504 and 12,288, respectively.

For each architecture, we report:
\begin{itemize}
\item \textbf{$\tau$ value},
\item \textbf{Test accuracy} (mean ± std.\ deviation in %),
\item \textbf{Training time} (mean ± std.\ deviation in minutes).
\end{itemize}

\subsection{Experimental Setup}
\begin{itemize}
\item Each configuration was run 5 times, and results are averaged.
\item Training time was measured on a single NVIDIA H200 GPU and includes model validation every 25 epochs (200 in total).
\item The entire training dataset was loaded at the beginning instead of subsequent batch loading.
\item Entries with ``--'' indicate configurations that were infeasible due to memory constraints.
\end{itemize}

\subsection{Key Observations}

Most of the observations reported for MNIST also apply to CIFAR-10. Therefore, here we focus only on additional aspects specific to this dataset.

\paragraph{Threshold count has limited effect}
The difference between using 4 or 7 thresholds is small, though the 7-threshold configuration shows a slight advantage in most cases. 

\paragraph{Training time is largest for L-type architectures}
For MNIST, L-type layers often trained faster than Top-K models, likely because GPU-accelerated dense matrix operations (\texttt{matmul}) are more efficient than sparse input selection used in Top-K. In CIFAR-10, however, L-type layers have the longest training times due to the much larger input size, which increases the number of connections in the first layer. Additionally, training time differences between the 4- and 7-threshold modes appear only for L-type architectures for the same reason; Top-K and F architectures are unaffected, as their number of connections does not depend on input size.

\paragraph{Accuracy standard deviation is larger than for MNIST}
Accuracy across 5 runs varies more than for MNIST, with standard deviations reaching up to 0.3\%. This is largely due to binarizing separate RGB channels with multiple thresholds, which increases variability in the input representation. Although CIFAR-10 has the same number of classes and slightly larger images than MNIST, this multi-threshold binarization amplifies sensitivity to initialization, data order, and training dynamics. Noise, labeling inconsistencies, and the small number of runs also contribute, whereas MNIST’s simple grayscale images with a single threshold yield highly stable results.

\paragraph{Best-performing architectures}
Across all settings, 1-layer L-type variants achieved the highest accuracy. For the largest budget architectures (e.g., 256K gates), memory constraints prevented evaluating L-type networks; in these cases, the best results were obtained using 2-layer Top-32 networks.

\begin{table*}[t!]
  \centering
  \begin{tabular}{@{}llrr rr rr@{}}
    \toprule
    & & & & \multicolumn{2}{c}{\textbf{7 thresholds}} & \multicolumn{2}{c}{\textbf{4 thresholds}} \\
    \cmidrule(lr){5-6} \cmidrule(lr){7-8}
    \textbf{\#gates} & \textbf{Model} & \makecell{\textbf{Layer} \\ \textbf{Width}} &
    \textbf{$\tau$}
      & \makecell{\textbf{Test Acc.} \\ {[\%]}} 
      & \makecell{\textbf{Train Time} \\ {[min]}} 
      & \makecell{\textbf{Test Acc.} \\ {[\%]}} 
      & \makecell{\textbf{Train Time} \\ {[min]}} \\
    \midrule
    \multirow{3}{*}{8,000} 
        & 1F      & 8,000 & 20  & 44.08 ± 0.12 & 2.5  ± 0.0 & 44.18 ± 0.27 & 1.3 ± 0.0 \\
        & 1Top32  & 8,000 & 20  & 52.27 ± 0.20 & 2.6  ± 0.0 & 51.86 ± 0.26 & 2.5 ± 0.0 \\
        & 1L      & 8,000  & 20 & \textbf{55.11} ± 0.17 & 45.4 ± 0.1 & 54.51 ± 0.25 & 26.1 ± 0.0 \\
    \midrule
    \multirow{5}{*}{64,000} 
        & 1F      & 64,000 & 90 & 49.17 ± 0.14 & 2.9  ± 0.0 & 49.06 ± 0.25 & 2.0 ± 0.1 \\
        & 1Top32  & 64,000 & 90 & 57.28 ± 0.30 & 15.3 ± 0.0 & 56.81 ± 0.19 & 16.9 ± 0.3 \\
        & 1L      & 64,000 & 90 & \textbf{57.66} ± 0.17 & 139.7 ± 0.1 & 57.64 ± 0.23 & 80.9 ± 0.1 \\
        & 2Top32  & 32,000 & 70 & 57.12 ± 0.11 & 72.9 ± 0.0 & 56.40 ± 0.17 & 72.7 ± 0.1 \\
        & 2L      & 32,000 & 70 & 55.75 ± 0.18 & 224.1 ± 0.1 & 55.58 ± 0.42 & 188.3 ± 0.4 \\
        
    \midrule
    \multirow{4}{*}{128,000} 
        & 1F      & 128,000 & 100 & 50.92 ± 0.24 & 3.8  ± 0.0 & 50.87 ± 0.30 & 3.7 ± 0.1 \\
        & 1Top32  & 128,000 & 100 & 59.43 ± 0.22 & 29.3 ± 0.0 & 58.95 ± 0.14 & 30.7 ± 0.1 \\
        & 1L      & 128,000 & 100 & --           & --          & 60.13 ± 0.14 & 157.3 ± 0.3 \\
         & 2Top32  & 64,000 & 90 & 58.88 ± 0.26 & 146.7 ± 0.0 & 58.36 ± 0.22 & 146.5 ± 0.0 \\
    \midrule
    \multirow{3}{*}{256,000} 
        & 1F      & 256,000 & 110 & 52.48 ± 0.28           & 5.0 ± 0.0          & 52.05 ± 0.09 & 5.3 ± 0.0 \\
        & 2F      & 128,000 & 100 & 54.76 ± 0.27 & 16.3 ± 0.0 & 54.18 ± 0.24 & 16.1 ± 0.0 \\
        & 2Top32  & 128,000 & 100 & \textbf{60.98} ± 0.19 & 297.2 ± 6.6 & 59.86 ± 0.23 & 298.7 ± 0.3 \\

    \bottomrule
  \end{tabular}
  \caption{Test accuracy and training time on CIFAR-10 for different logic gate networks, layer widths, depths and input binarization thresholds. Each cell shows the mean ± standard deviation over 5 runs. Dashes indicate inapplicable configurations. Bolded values indicate the chosen models (LILogic Net-S, -M, -L) with best performance.}
  \label{tab:cifar_combined}
\end{table*}

\section{Training Accuracy and Effect of Discretization}

We analyze the influence of discretization on network performance by measuring the difference between the accuracy in inference mode and the accuracy during differentiable training. Early in training, this binarization effect can noticeably affect accuracy as the network starts by learning a probabilistic, differentiable LGN, with high uncertainty in the selection of individual logic gates. The discretization step—selecting the most probable gate for inference—can therefore produce substantial changes in network behavior during this phase.

\begin{figure}[ht]
    \centering
    \includegraphics[width=\linewidth]{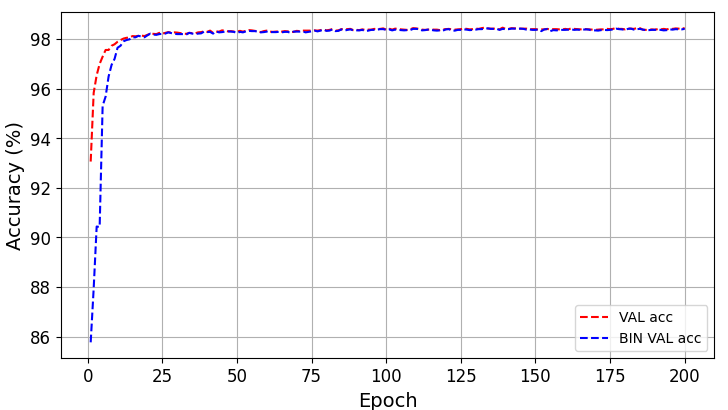}
    \caption{MNIST LILogic Net-S training and validation accuracy curves before (VAL acc) and after (BIN VAL acc) discretization. The
discretization error is very small during late training.}
    \label{fig:bin_err}
\end{figure}

As training progresses, the gate selection probabilities become more confident, and the effect of discretization on accuracy diminishes (e.g. Fig.~\ref{fig:bin_err}). By the end of training, the difference between inference-mode and training-mode accuracy is negligible, with a final binarization error \textbf{below 0.1 percentage points}.

This pattern demonstrates that the network first explores a smooth, probabilistic logic representation and gradually converges to stable discrete logic choices, with minimal impact on final performance.

\section{LGN Model FPGA Resource Usage}

Table~\ref{tab:FPGA1} summarizes the accuracy and FPGA resource utilization of different LGN configurations across MNIST, FashionMNIST, and CIFAR-10 datasets, including LUT usage for generic and target devices.

Tables~\ref{tab:FPGA1} and ~\ref{tab:FPGA2} summarize the hardware characteristics of various LGN configurations across MNIST, FashionMNIST, and CIFAR-10 datasets. Table~\ref{tab:FPGA1} reports accuracy and FPGA resource utilization (LUT usage for generic and Xilinx devices), while Table~\ref{tab:FPGA2} reports Xilinx frequency and logic/routing latency. Architectures marked with $*$ correspond to DiffLogicNet variants. Dashes (-) indicate missing or not applicable data.

\begin{table*}[t!]
\centering
\caption{Accuracy of single runs and FPGA resources per dataset. Most of the achitectures are binarized LILogicNet-S,-M,-L. \\ $*$ Architectures of DiffLogicNet-S and DiffLogiCNet reported for MNIST. Given two values of LUTs means total LUTS and LUTs without population count (popcount) part.}
\begin{tabular}{lcccccc}
\toprule
\textbf{Dataset} & \textbf{Architecture} & \textbf{Acc [\%]} & \textbf{Generic LUT4} & \textbf{ICE40 LUT4} & \textbf{Generic LUT6} & \textbf{Xilinx LUT6} \\
\midrule
\multirow{4}{*}{MNIST} 
    & 1x4000   & 97.63 & 11987 / 3637  & 11188 & 8760 / 3637   & 7103 / 1791 \\
    & 1x8000   & 98.47 & 21677 / 7047  & 20696 & 16811 / 7047  & 14076 / 3491 \\
    & 2x16000  & 98.94 & 47869 / 14940 & 46379 & 41258 / 14940 & 37373 / 11833 \\
    & 4x8000   & 98.79 & 24676 / 8681  & 23697 & 20806 / 8065  & 16004 / 4637 \\
\midrule
\multirow{4}{*}{FMNIST} 
    & 1x8000        & 90.03 & 32436 / 7695   & 27163 & -           & 14321 / 3736 \\
    & 2x16000       & 90.03 & 51311 / 14891  & 50893 & -           & 37075 / 11535 \\
    & 2x32000       & 90.39 & 95342 / 29111  & 93926 & -           & 73938 / 22308 \\
    & 2x64000       & 90.66 & 176882 / 55302 & 172196 & -          & 143226 / 39769 \\
    & 6x8000* & 88.56 & 39347 / 17562 & 39417 & -          & 22710 / 10377 \\
    & 6x64000* & 90.27 & 264873 / 131992 & 264982 & -      & 172787 / 72807 \\
\midrule
\multirow{3}{*}{CIFAR10} 
    & 1x8000     & 54.89 & 43237 / 7834  & -     & 38225 / 7834  & 14415 / 3830 \\
    & 1x64000    & 58.06 & 186010 / 57329 & 187468 & -           & 104853 / 27070 \\
    & 2x128000   & 60.76 & 387156 / 117164 & -    & 339376 / 117164 & 293285 / 92353 \\
\bottomrule
\end{tabular}
\label{tab:FPGA1}
\end{table*}

\begin{table*}[t!]
\centering
\caption{Xilinx XC7A200T frequency and latency per dataset and model architecture. Most of the achitectures are binarized LILogicNet-S,-M,-L. $*$ Architectures of DiffLogicNet-S and DiffLogicNet reported for MNIST. Dashes (-) indicate missing or not applicable data.}
\begin{tabular}{lcccc}
\toprule
\textbf{Dataset} & \textbf{LGN} & \textbf{Xilinx Freq} & \textbf{Logic Latency (ns)} & \textbf{Routing Latency (ns)} \\
\midrule
\multirow{4}{*}{MNIST} 
    & 1x4000   & 40.15 MHz & 5.1  & 19.8 \\
    & 1x8000   & 32.89 MHz & 6.0  & 24.4 \\
    & 2x16000  & 29.71 MHz & 6.2  & 27.4 \\
    & 4x8000   & 26.69 MHz & 5.0  & 32.4 \\
\midrule
\multirow{6}{*}{FMNIST} 
    & 1x8000        & 33.76 MHz & 5.9  & 23.8 \\
    & 2x16000       & 20.13 MHz & 6.1  & 43.6 \\
    & 2x32000       & 18.16 MHz & -    & -    \\
    & 2x64000       & -         & -    & -    \\
    & 6x8000* & 19.11 MHz & - & - \\
    & 6x64000* & -       & - & - \\
\midrule
\multirow{3}{*}{CIFAR10} 
    & 1x8000     & 31.18 MHz & 5.8  & 26.2 \\
    & 1x64000    & -         & -    & -    \\
    & 2x128000   & -         & -    & -    \\
\bottomrule
\end{tabular}
\label{tab:FPGA2}
\end{table*}

% \section{Rationale}
% \label{sec:rationale}
% % 
% Having the supplementary compiled together with the main paper means that:
% % 
% \begin{itemize}
% \item The supplementary can back-reference sections of the main paper, for example, we can refer to \cref{sec:intro};
% \item The main paper can forward reference sub-sections within the supplementary explicitly (e.g. referring to a particular experiment); 
% \item When submitted to arXiv, the supplementary will already included at the end of the paper.
% \end{itemize}
% % 
% To split the supplementary pages from the main paper, you can use \href{https://support.apple.com/en-ca/guide/preview/prvw11793/mac#:~:text=Delete%20a%20page%20from%20a,or%20choose%20Edit%20%3E%20Delete).}{Preview (on macOS)}, \href{https://www.adobe.com/acrobat/how-to/delete-pages-from-pdf.html#:~:text=Choose%20%E2%80%9CTools%E2%80%9D%20%3E%20%E2%80%9COrganize,or%20pages%20from%20the%20file.}{Adobe Acrobat} (on all OSs), as well as \href{https://superuser.com/questions/517986/is-it-possible-to-delete-some-pages-of-a-pdf-document}{command line tools}.